\documentclass[times,twocolumn,final,authoryear]{elsarticle}
\usepackage{prletters}
\usepackage{hyperref}

\usepackage{graphicx}
\usepackage{subfigure}

\usepackage{amssymb}

\usepackage[nodots]{numcompress}

\usepackage{tabularx}

\newcommand{\ie}[0]{{\em i.e.}}
\newcommand{\eg}[0]{{\em e.g.}}
\newcommand{\Xfig}[1]{Fig.~\ref{#1}}
\newcommand{\Xfigs}[1]{Figs.~\ref{#1}}
\newcommand{\Xsec}[1]{Section~\ref{#1}}

\usepackage{color}

\newcommand{\ELIMINE}[1]{}

\begin{document}




\title{A graph-based mathematical morphology reader}


\author{Laurent Najman}
\author{Jean Cousty}
\address{Universit\'e Paris-Est, Laboratoire d'Informatique
  Gaspard-Monge, \'Equipe A3SI, ESIEE Paris.\\
E-mail: \{laurent.najman, jean.cousty\}@esiee.fr\\[0.5cm]
The two authors contributed equally to this work which received funding from the Agence Nationale de la Recherche, contract
ANR-2010-BLAN-0205-03.
}
\begin{abstract}
  This survey paper aims at providing a ``literary'' anthology of
  mathematical morphology on graphs. It describes in the English
  language many ideas stemming from a large number of different papers,
  hence providing a unified view of an active and diverse field of
  research.
\end{abstract}


\maketitle


\section{Introduction}

Mathematical morphology was born almost 50 years ago~\citep{Serra-82},
initialy an evolution of a continuous probabilistic
framework~\citep{matheron75}. Historically, this was the first
consistent non-linear image analysis theory which from the very start
included not only theoretical results, but also many practical
aspects, including algorithmic ones~\citep{Soille-1999}. Despite its
continuous origin, it was soon recognized that the roots of this
theory were in algebraic theory, notably the framework of complete
lattices~\citep{heijmans94book}. This allows the theory to be
completely adaptable to non-continuous spaces, such as graphs. For a
survey of the state of the art in mathematical morphology, we
recommend~\citep{NajTal10}.

Graphs are generic data structures that have a long history in
mathematics and have been applied in almost every scientific and
engineering field, notably image analysis and computer
vision~\citep{Book-Grady-Lezoray--2012,grady2010discrete}. Because of
their many interesting properties, a current trend is to develop the
classical continuous tools from signal processing onto this kind of
structures~\citep{shuman2013emerging}.

The usefulness of graphs for mathematical morphology has long been
recognized~\citep{Vincent-89}, and the same trend as in the signal
processing community can be observed
here~\citep{najman-meyer-GraphBook}. The objective of this paper is to
offer an overview of the advantages of graphs for mathematical
morphology. To reach a wider audience, we decided to express all the
ideas with the least possible mathematical jargon, if possible without
any equation whatsoever. We emphasize that point by using the word
{\em reader} in the title. This paper aims at being a ``literary''
anthology of papers using graph in the field of mathematical
morphology, describing in the English language the main ideas of many
papers, pointing out where the interested researcher can find more
details.

This paper is organized as follows. Section~\ref{sec:graphs} describes
what is a graph, what type of graphs can be encountered, and how we
can build them. Section~\ref{sec:adjunc} explains the basis of
algebraic morphology and what are the adjunctions that are used on
graphs for defining elementary morphological operators. One of the
most basic problem in graphs is finding paths, and
section~\ref{sec:paths} gives an overview of what has been done with
paths in the field. The next section~\ref{sec:conwat} is divided in
three parts. In the first part (section~\ref{sec:segwat}), two major
morphological tools for segmentation, namely the watershed and the
flat zone approach, are reviewed. The second part
(section~\ref{sec:confilt}) deals with their close cousin, connective
filtering. Combining these two parts togethers provide hierarchical
segmentation and filtering, which is the object of
section~\ref{sec:hiermst}. Section~\ref{sec:further} exhibits some
links between graph-morphology and discrete calculus. Before
concluding the paper, a penultimate section~\ref{sec:beyond} describes
several interesting structures that generalize graphs.

\section{What is a graph and some examples of graphs for morphological
  processing}
\label{sec:graphs}

A graph is a representation of a set of data where some pairs of data
are connected by links. Once a graph representation is adopted, the
(abstraction of) interconnected data are called vertices or nodes of
the graph and the links that connect vertices are called edges. An
edge of the graph is then simply a pair of connected vertices. Thus, a
graph is made of a set of vertices and of a set of edges. If needed,
we can also associate to each vertex and/or to each edge a weight that
represents some kind of measure on the data, leading to weighted
graphs. Once a graph is specified, the neighbors of a data point can
be obtained by considering the edges that link this data point to
others in the graph. Conversely, if we know the neighbors of each data
point, then we can obtain edges by considering all pairs of
neighbors. Thus, another common (and equivalent) way to define a graph
is to consider the sets of neighbors of each vertex instead of a set
of edges; in this case, the neighbourhood relation is symmetrical.

\begin{figure*}
    \begin{center}
    \begin{tabular*}{1\linewidth}{@{\extracolsep{\fill}}c c c c c}
      \includegraphics[width=0.18\linewidth]{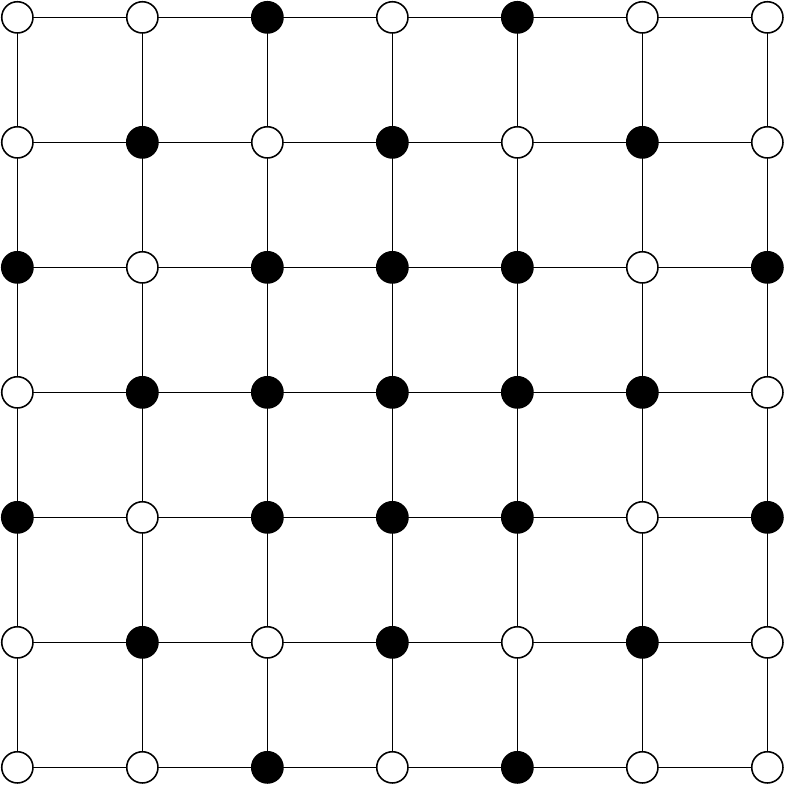}&
      \includegraphics[width=0.18\linewidth]{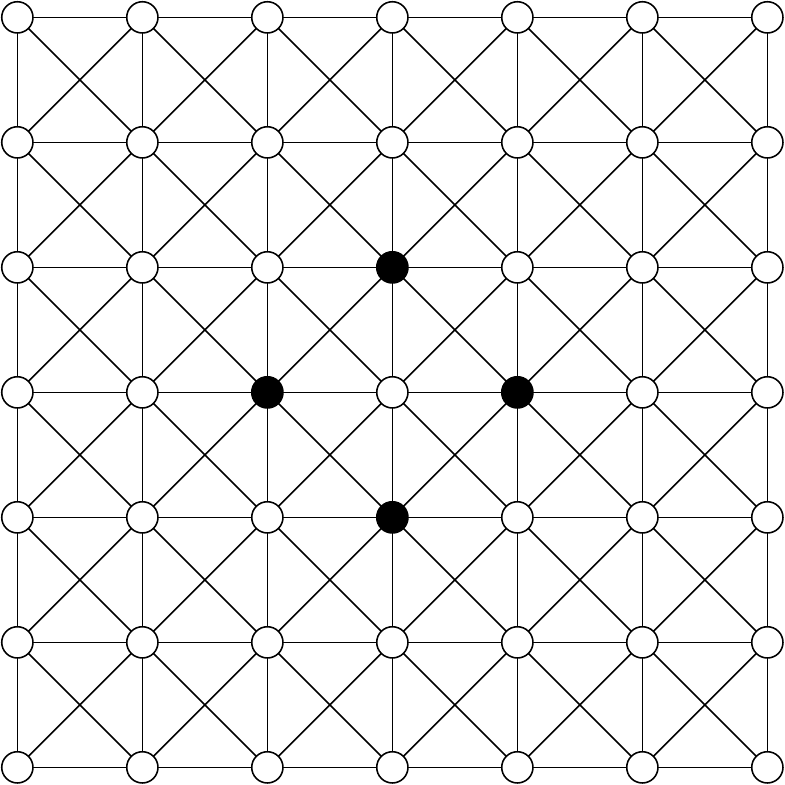}&
      \includegraphics[width=0.18\linewidth]{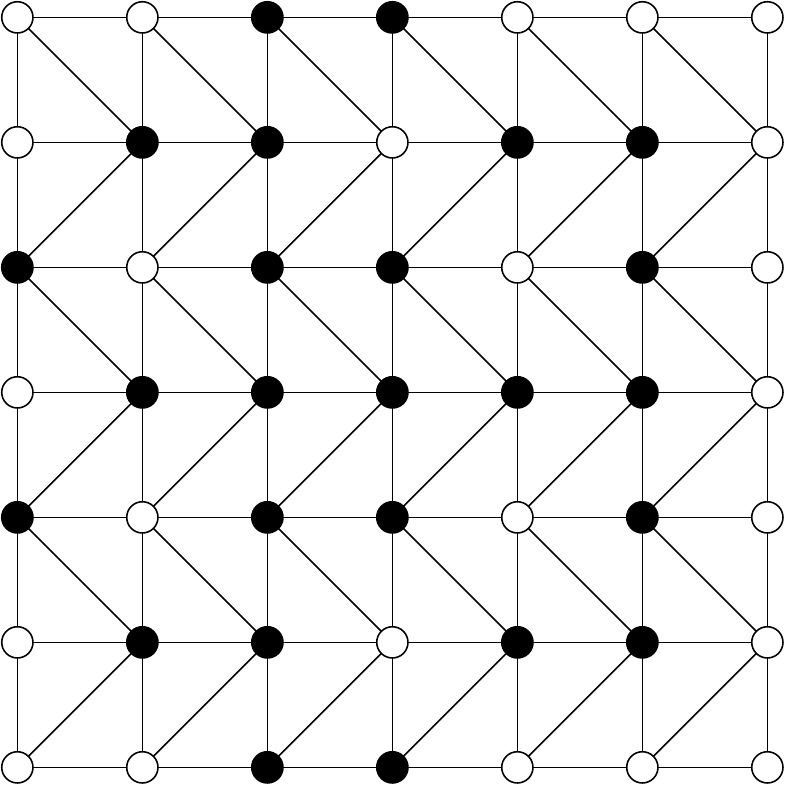}&
      \includegraphics[width=0.18\linewidth]{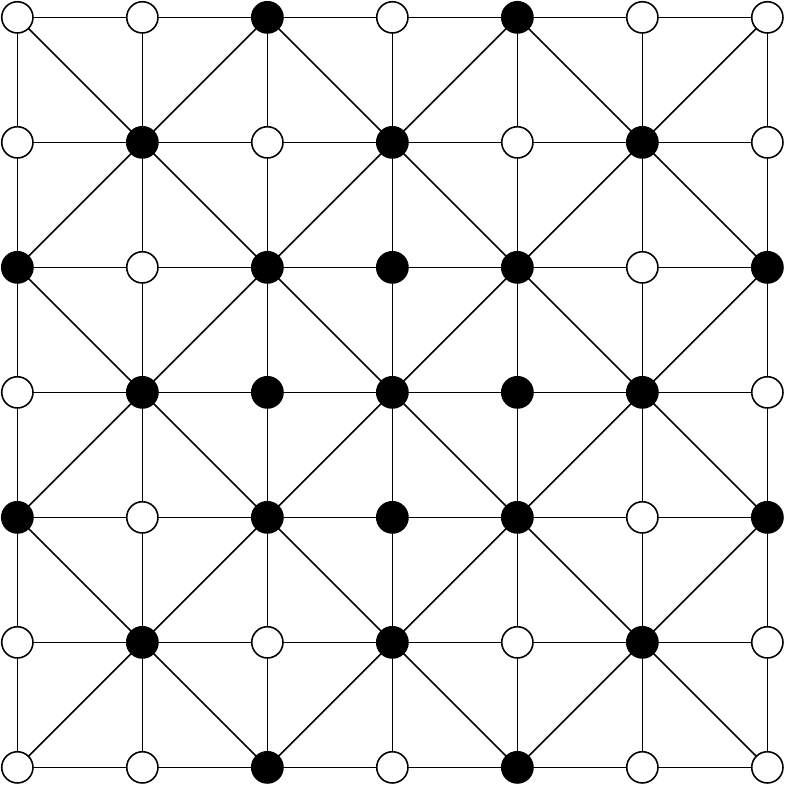}&
      \includegraphics[width=0.18\linewidth]{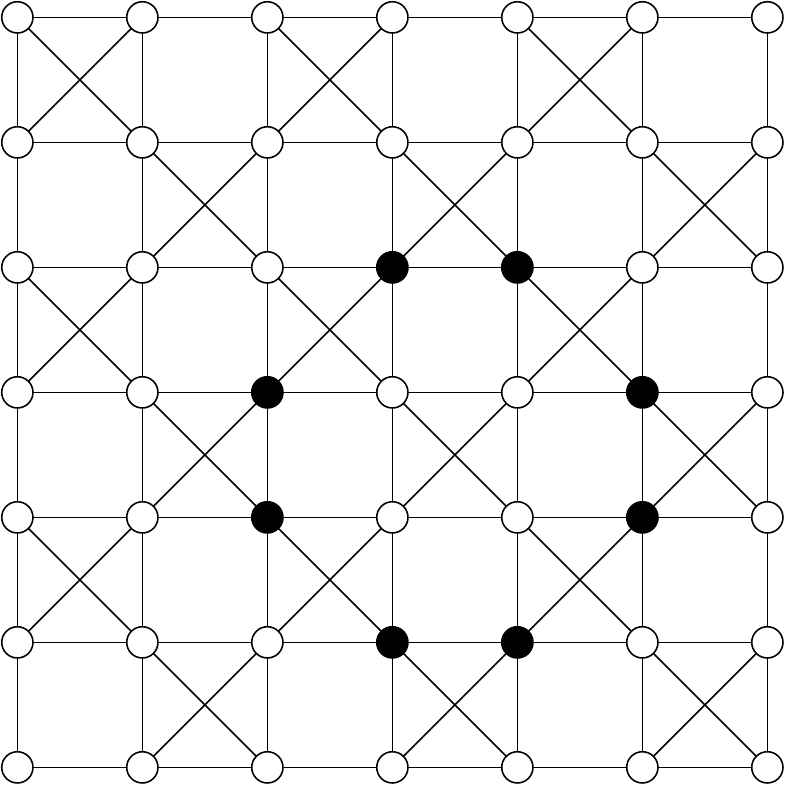}\\
      (a) 4-adjacency grid & (b) 8-adjacency grid & (c) 6-adjacency grid &
      (d) Khalimsky grid & (e) Perfect fusion grid 
\end{tabular*}
\caption{\label{fig:grids} Examples of pixel adjacency graphs used in
  image processing. Vertices are represented by dots and edges are
  represented by line segments. In sub-figures (b) and (e), the black
  vertices form the discrete analog of a Jordan curve but they do not
  separate the white vertices into two connected regions. In
  sub-figures (a,c,d), the white vertices form regions separated by
  frontiers made of black vertices; the frontiers are thick but they
  cannot be further ``thinned'' or ``reduced by black point removal''
  while leaving unchanged the number of white regions. }
  \end{center}
\end{figure*}

In image processing, the first (historically) example is the case of
an image itself: indeed, an image is a set of pixels with integer
coordinates and color information. These pixels are often structured
in a grid thanks to the classical pixel adjacency relation (\ie, 4- or
8- adjacency in 2D \citep{KR-89}, see \Xfigs{fig:grids}(a) and
(b)). For example, a pixel is connected to the 4 or 8 closest pixels
according to the Euclidean distance between the integer
coordinates. In the associated graph representation, pixels are
vertices, and if two pixels are connected for the given
grid-adjacency, they are linked by an edge of the graph. In the
literature, the set of vertices is often denoted by~$V$ (for vertices)
or~$N$ (for nodes); the set of edges is generally denoted by~$E$. The
weight of a vertex can be as simple as the gray value of the
corresponding pixel, or as complex as a measure combining color
information and other cues, {\em etc.}, taken on a patch around the
pixel. The weight of an edge is generally a kind of distance between
the data of the pixels linked by the given edge. For example, in the
case of a gray-scale image, the edge weight can be a gradient of
intensity such as the absolute difference between pixel intensities.

Some important topological properties cannot be recovered when only
the 4- (or the 8-) adjacency graph is considered \citep{KR-89}. The
Jordan curve theorem, which states that a closed curve separates the
2D space into two regions (interior and exterior) does not hold true
in this setting (see \eg{} \Xfig{fig:grids}(b)). This has lead
researchers to explore other adjacency relations \citep{AGL-96} such
as the 6-adjacency grid (also known as the hexagonal grid, see \eg{}
\citep[Chapter VI]{Serra-82} and \Xfig{fig:grids}(c)) or grids derived
from the Khalimsky plane \citep{KKM-90} (see \Xfig{fig:grids}(d)) for
which a discrete analog of the Jordan curve theorem can be
expressed. This, as well as better isotropic properties, explains the
popularity of the hexagonal grid for morphological
processing. However, in contrast to other grids, the hexagonal grid
cannot be easily extended to 3D or higher dimensional spaces (see
\eg{} \citep{SS-06}). Another problem, which can be encountered with
any of the 4-, 6- or 8- adjacency grid, is related to the thickness of
frontiers or contours made of vertices: a contour can contain an
arbitrary number of interior points (\ie{} points in the contour that
are not adjacent to the complement of the contour)
\citep{CBCN-08,CCNB-08}. With the perfect fusion grid (see
\Xfig{fig:grids}(e)) studied in \citep{CB-09} a contour is always
thin. This thinness property of contours is related (by an equivalence
theorem) to an interesting properties dealing with the merging of
adjacent regions \citep{CBCN-08}. This latter property, which is
indeed satisfied in perfect fusion grids, gave its name to this
adjacency relation.

The graphs obtained with the adjacency relations presented in the
previous paragraphs are ``regular''. For instance, with the
4-adjacency relation, each vertex has 4 neighbors and the patterns
given by the neighborhoods of the vertices are all the same. Thus, the
graph is invariant under translation, {\em i.e.}
if one translates the original pixel coordinates, then one still
obtains the same graph\footnote{According to \citet{Burhardt2001}, one
  should say {\em equivariant} when the operator commutes with
  translation}.  The operator acting on the images through this kind
of graphs are then called spatially invariant and were historically
the first ones to be considered in mathematical morphology. Since
2005, spatially variant morphology has become increasingly
interesting~\citep{LDM-05,LDM-07}. The idea is to adapt the local
configuration around a point to the image content: a pixel is no more
adjacent to its 4- or 8-neighbors but to a pattern that locally
corresponds to the image content. The local patterns can be obtained
by removing some edges of a spatially invariant graph. In this case,
one can threshold some edge weights to determine the edges that are
kept (see \eg{} \citep{CNDS-13}). One can also apply a non-local
selection procedure such as keeping only the edges of a minimum
spanning tree (whose definition is given in \Xsec{sec:segwat}) of the
initial graph \citep{SM-09}. In these cases one obtains a graph that
has less edges than the initial spatially invariant graph. It can also
be interesting to have more edges or to connect pixels whose
coordinates are far from each other. To this end, one may find, for
each pixel of the image the closest pixels for some distance that is
not only based on the coordinates. The distance can be a geodesic
distance in a weighted graph (see more details in the next
\Xsec{sec:paths}) or can be a distance related to a continuous feature
space onto which the vertices are mapped. Therefore, the distance
between two pixels with the same color can be low even if the pixels
are localized far from each other. Then, the neighbors of a pixel can
be all pixels at a distance less than a predefined value
\citep{LDM-05,CLB-12} or can be the~$k$ closest pixels for the chosen
distance, where~$k$ is a predefined value that sets up the size of the
neighborhood in the resulting graph (see \eg{} \citep{FH-04}). In the
framework of second generation connectivity
\citep{Serra-88,Ronse-95,heijmans1999connected}, operators from
mathematical morphology itself have been used \citep{OW-07} to
determine the pairs of (long-distance) nodes that should be
connected. Finally, in the framework of non-local means image
filtering \citep{BBM-05}, a complete graph is considered to structure
the image pixels (\ie{}, any two pixels are linked by an edge). Each
edge is then weighted by a similarity measure between small patches
centered in the pixels corresponding to the extremities of the given
edge.

Apart from regular grids, one of the first kind of graphs used for
morphological processing was probably the family of region adjacency
graphs \citep{Pavlidis,Vincent-89,Beucher-94,Meyer-94}. The nodes
of the graph, often called super-pixels, are the faces of a
tessellation (or, using words from the digital world, the regions of a
segmentation) of the space. Two faces are linked by an edge if they
are neighbor of each other for a certain predefined adjacency. In
mathematical morphology, the faces are often obtained as the flat
zones of an image or the catchment basins of a watershed of the
gradient magnitude of the image (see \Xsec{sec:segwat} for more
details on these methods). However, any pixel classification method
can lead to such a region adjacency graph.

\begin{figure*}[htb]
  \begin{center}
    \begin{tabular}{m{0.3\linewidth} m{0.3\linewidth} m{0.3\linewidth}}
      \begin{center}\includegraphics[height=0.9\linewidth]{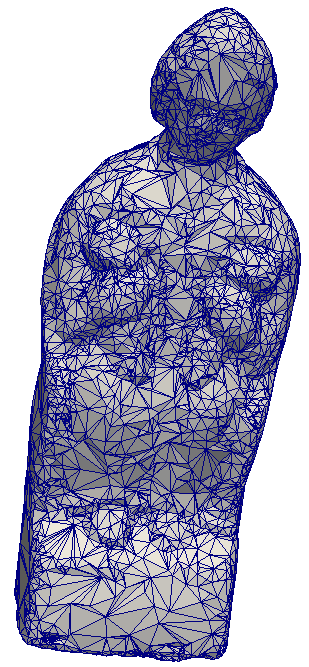}\end{center}&
      \begin{center}\includegraphics[height=0.7\linewidth]{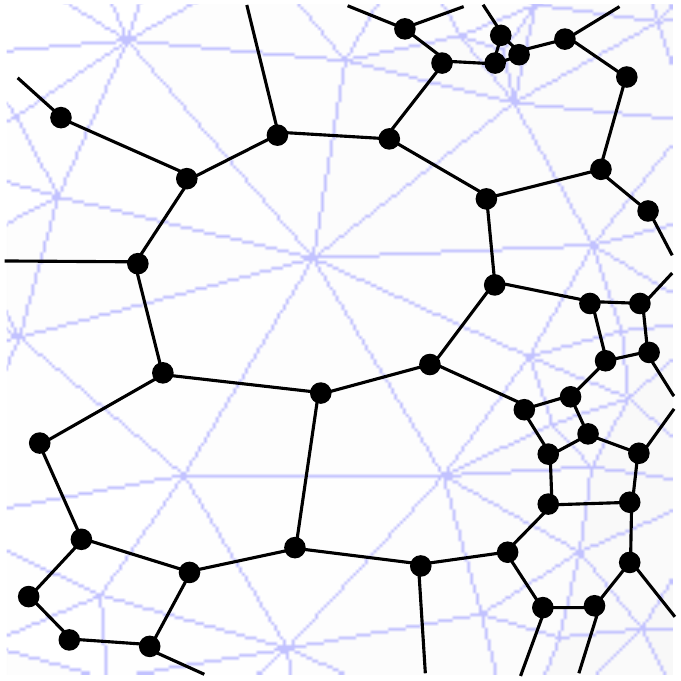}\end{center}&
      \begin{center}\includegraphics[height=0.9\linewidth]{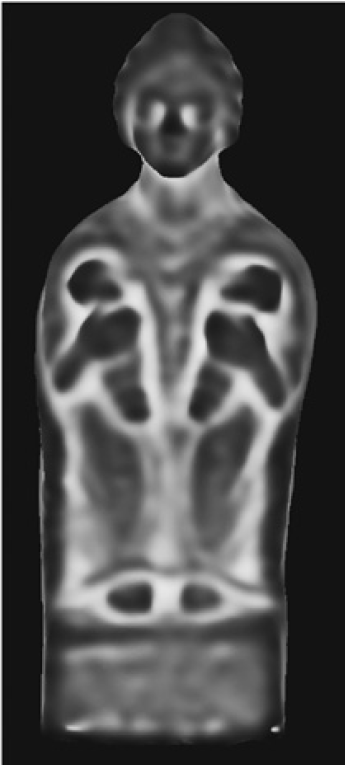}\end{center}\\
      \end{tabular}
    \begin{tabular}{m{0.3\linewidth} m{0.3\linewidth} }
      \begin{center}\includegraphics[height=1.4\linewidth]{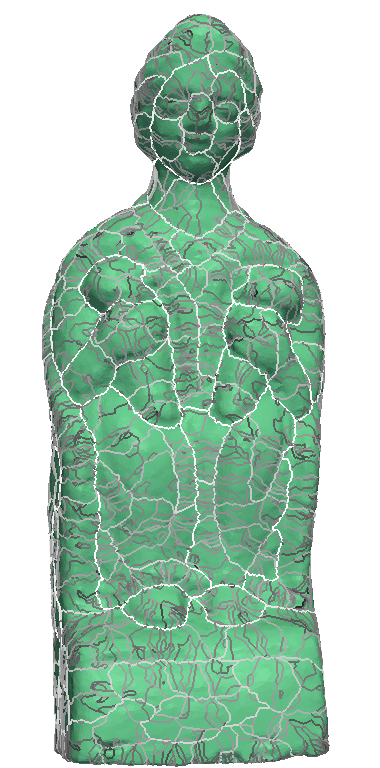}\end{center}&
      \begin{center}\includegraphics[height=1.2\linewidth]{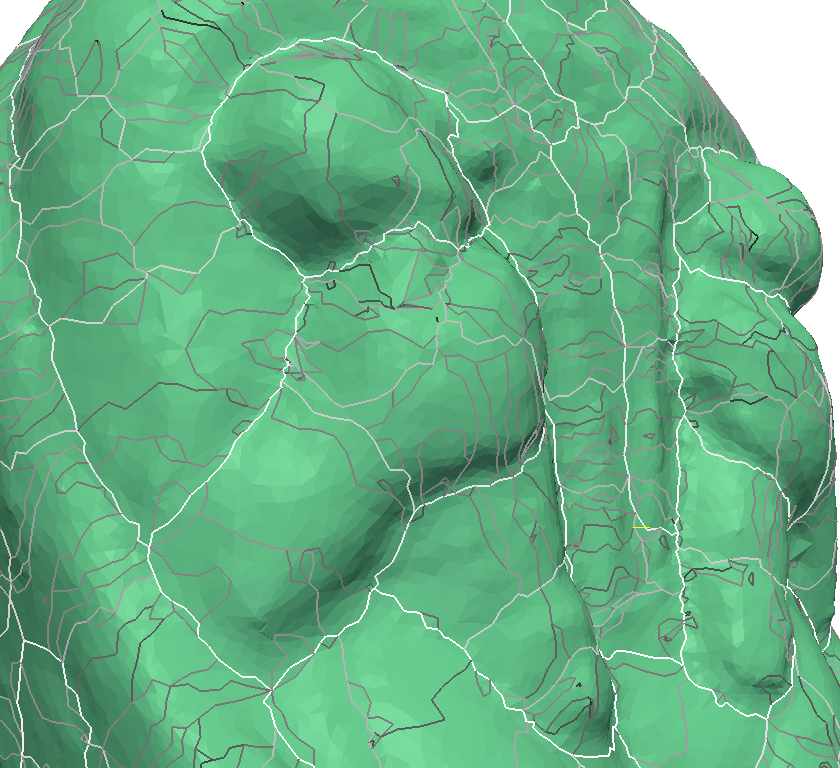}\end{center}\\
    \end{tabular}
    \caption{\label{fig:statues} Illustration of the segmentation of
      the surface of a 3D object. First row: a triangular mesh, a crop
      on its associated dual graph, and its pseudo-inverse
      curvature. Second row: a saliency map representing a
      hierarchical segmentation of the surface. A framework for the
      indexing and retrieval of ancient artwork 3D models, using shape
      descriptors adapted to the surface regions of the segmentations,
      is detailed in \citep{PJNC-11}. The mesh is provided by the
      French Museum Center for Research and Restoration (C2RMF, Le
      Louvre, Paris).}
  \end{center}
\end{figure*}

Besides image analysis, graphs are often used in computer
graphics. Indeed, a triangular mesh (or a triangulation), which is a
very common representation for the surface of a 3D object, can be
processed as a graph (see \eg{}, \Xfig{fig:statues}). A triangular
mesh is composed of triangles, sides (line segments) and corners
(points) glued together according to certain rules (\eg{} two
triangles can have a common side or a common corner). Given a
triangular mesh one can consider the graph whose vertices are the
corners of the triangles and whose edges are the pairs of corners that
are the extremities of a same side. When the triangular mesh satisfies
the additional rule of a pseudomanifold (\ie{} when each side belongs
to exactly two triangles), a dual graph can also be built: each
triangle is a vertex of the graph and two vertices are linked by an
edge if the corresponding triangles share a common side. The vertices
and edges of these graphs can be weighted with an information relative
to the mesh: this can be a colorimetric or a geometric
information. For instance, it is possible to weight these graphs with
a function related to the curvature of the surface
\citep{MWR-99,PJNC-11}. Another possibility is to weight each edge of
the dual graph with the face angle between the corresponding two
triangles.

In computer graphics, unstructured cloud points are also often
available. In order to build a graph over this data, one can again
consider the closest neighbors of each point for a given
distance. Another interesting possibility consists of building the
Delaunay triangulation of the cloud points and to derive a graph from
this triangulation. In the computer graphics community, this leads to
a multiscale hierarchical representation of the data, called the
$\alpha$-shapes \citep{EM-94}, that were later considered in morphology
by \citet{LS-08,LR-12}.

To finish this section, let us mention two applications of
mathematical morphology in original graphs. In the first one,
morphological segmentation operators are used as an image classifier
\citep{PFAT-12}. To this end, a given image database is structured by a
weighted graph before applying morphological operators: each vertex is
an image and two related images are connected by an edge that is
weighted by a similarity measure. In the second
application~\citep{XGN-12}, graph based morphology is used for
regularizing the features associated to a shape space representing an
image. The shape space is a weighted graph called the component tree
of the image (see more details in \Xsec{sec:confilt}). The nodes are
the shapes (components) appearing in the images and there is an edge
between two shapes if they are included in each other. The weight of
the nodes are provided by the shape descriptors.

\section{Adjunctions and basic morphological operators}
\label{sec:adjunc}
The algebraic basis of mathematical morphology is the lattice
structure and the morphological operators act on lattices
\citep{Serra-88,HR-90,RS-10}. In other words, the morphological
operators map the elements of a first lattice to the elements of
second one (which is not always the same as the first one). A lattice
is a partially ordered set such that for any family of elements, we
can always find a least upper bound and a greatest lower bound (called
a supremum and an infimum). The supremum ({\em resp.}, infimum) of a
family of elements is then the smallest (greatest) element among all
elements greater (smaller) than every element in the considered
familly.

The classical lattice for binary image processing contains all shapes
which can be drawn in the considered image, namely it is the family of
all subsets of image pixels. The supremum is given by the union and
the infimum by the intersection. A morphological operator is then a
mapping that associates to any subset of pixels (a shape) another
subset of pixels. Similarly, given a graph, one can consider the
lattice of all subsets of vertices \citep{Vincent-89} and the lattice
of all subsets of edges \citep{CNS-09,CNDS-13}. The supremum and
infimum in these lattices are also the union and intersection.  In
some cases, it also interesting to consider a lattice whose elements
are graphs, so that the inputs and outputs of the operators are
graphs. In particular, when the workspace is a graph (\eg{} a pixel
adjacency graph defined from an image), it is interesting to consider
the lattice of all its subgraphs~\citep{CNS-09,CNDS-13}: a graph is a
subgraph of another when both the vertex and edge sets of the two
graphs are included in each other. In the lattice of subgraphs, the
supremum or union ({\em resp.}, the infimum or intersection) of two graphs is
defined by the union ({\em resp.}, intersection) of the vertex and edge sets.

\begin{figure*}[htb]
  \begin{center}
    \begin{tabular*}{0.95\linewidth}{@{\extracolsep{\fill}}c c }
   \includegraphics[width=0.43\linewidth]{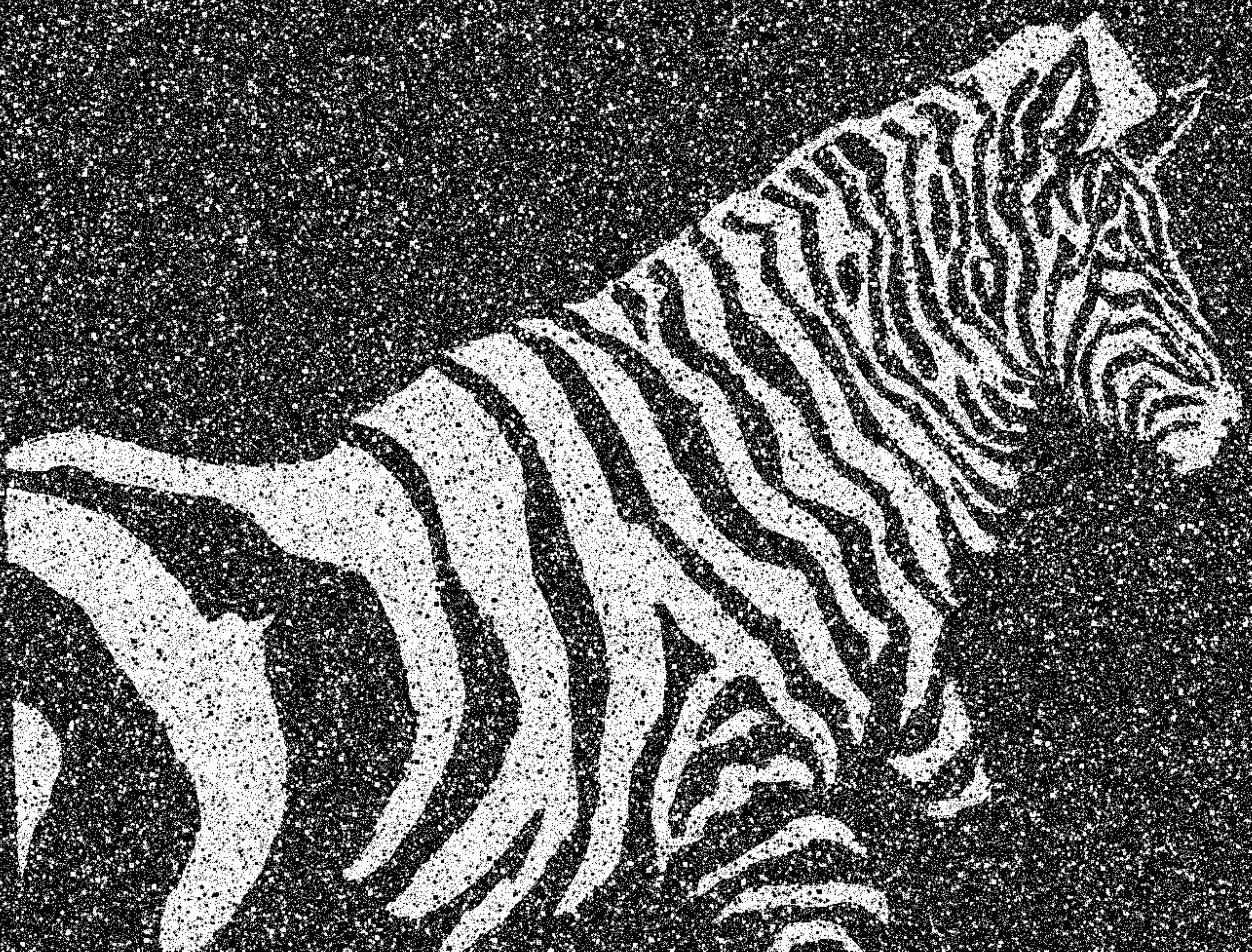}&
   \includegraphics[width=0.43\linewidth]{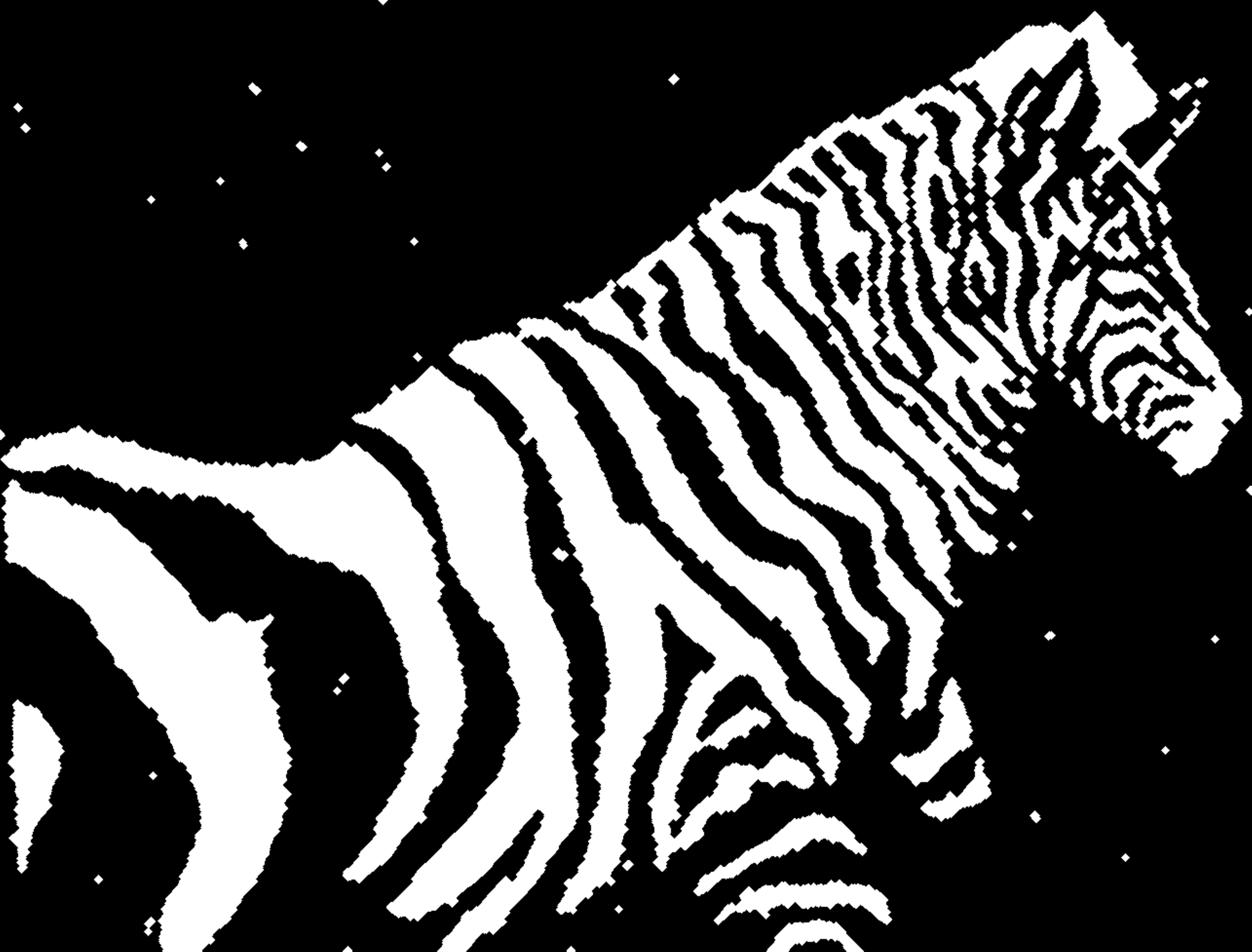}\\[0.2cm]
    \includegraphics[width=0.43\linewidth]{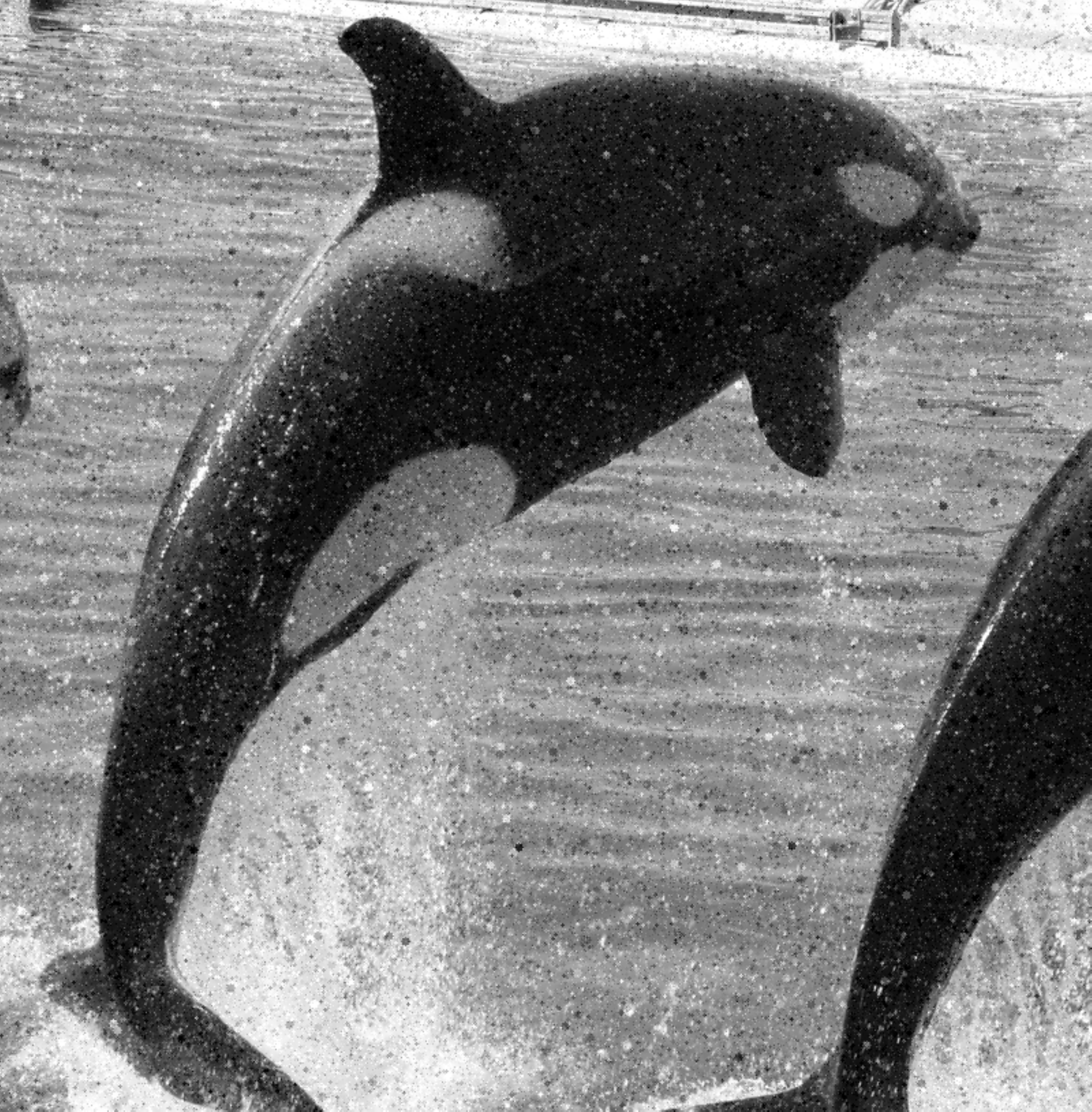}&
    \includegraphics[width=0.43\linewidth]{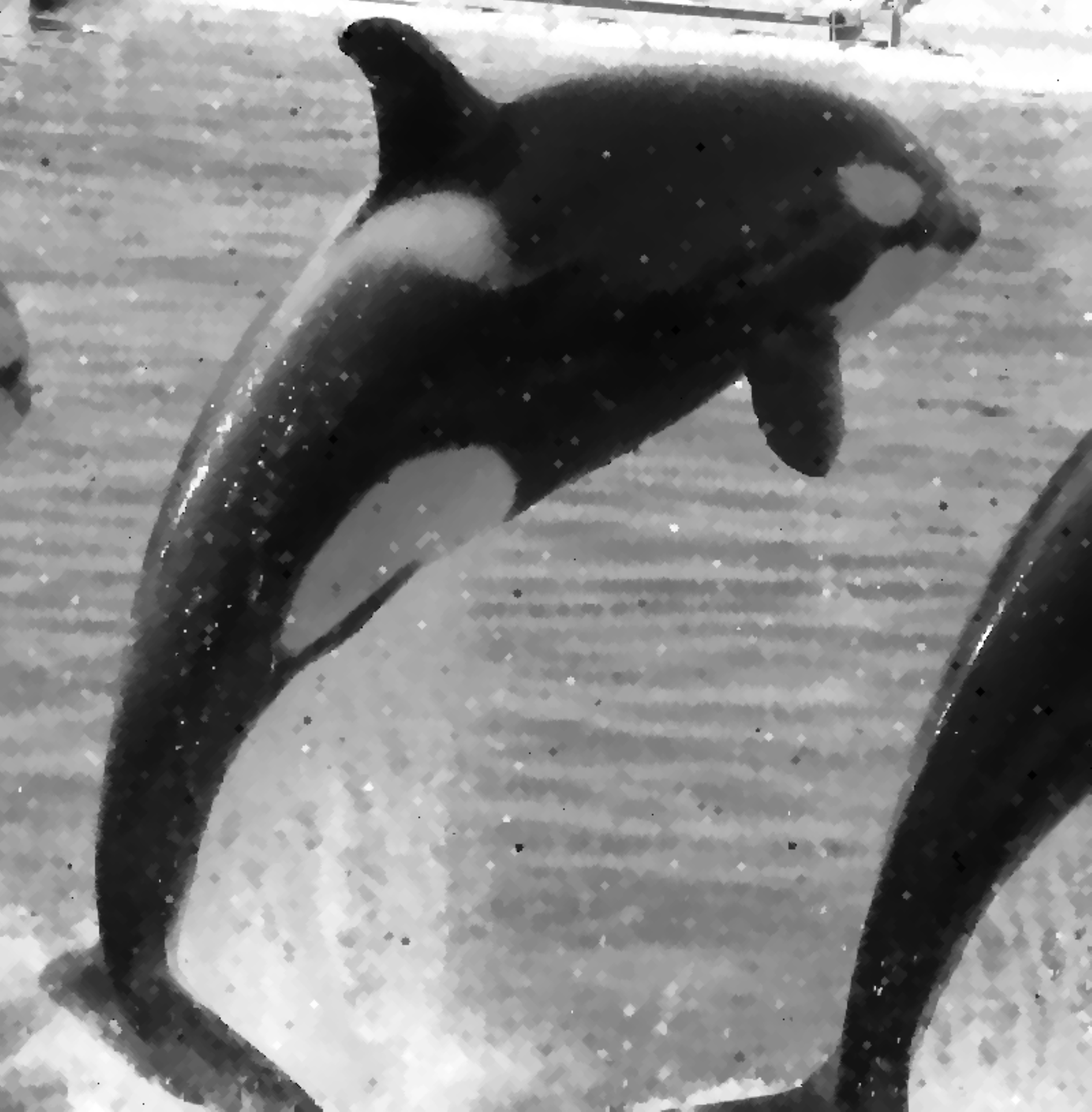}
    \end{tabular*}
    \caption{\label{fig:IllusASF} Illustration of morphological
      alternating sequential filters on graphs. The alternating sequential
      filters are obtained thanks to the vertex-edge and edge-vertex
      dilations presented in \Xsec{sec:adjunc}. Top ({\em resp.},
      bottom) row: the filtering (right) is applied to a binary ({\em
        resp.}, grayscale) image (left) considered in the 4-adjacency
      graphs ({\em resp.}, in a spatially variant adjacency
      graph). The corresponding filterings in the usual pixel-based
      framework of structuring elements (\ie{} the filters obtained on
      graphs from the natural dilation) are less performing (see
      details in \citep{CNDS-13}).}
  \end{center}
\end{figure*}

The algebraic framework of morphology relies mostly on a relation
between operators called adjunction \citep{Serra-88,HR-90}. This
relation is particularly interesting, because it extends single
operators to a whole family of other interesting operators: having a
dilation ({\em resp.}, an erosion), an (adjunct) erosion ({\em resp.},
a dilation) can always be derived, then by applying successively these
two adjunct operators a closing and an opening are obtained in turn
(depending which of the two operators is first applied), and finally
composing this opening and closing leads to alternating filters. Each
of these operators satisfy a set of remarkable properties that are
interesting in particular in the context of noise cleaning (more
details on the use of morphological operators for image denoising are
provided in the next paragraphs and illustrated in
\Xfig{fig:IllusASF}). Firstly, they are all increasing, meaning that
if we have two ordered elements, then the results of the operator
applied to these elements are also ordered, so the morphological
operators preserve order. Additionally the following important
properties hold true:
\begin{itemize}
  \item the dilation ({\em resp.}, erosion) commutes under supremum
    ({\em resp.}, infimum);
  \item the opening, closing and alternating filters are indeed
    morphological filters, which means that they are both increasing
    and idempotent (after applying a filter to an element of the
    lattice, applying it again does not change the result);
  \item the closing ({\em resp.}, opening) is extensive
    ({\em resp.}, anti-extensive), which means that the result of the
    operator is always larger ({\em resp.}, smaller) than the initial object;
\end{itemize}

In binary morphology on a graph, as initially proposed by
\citet{Vincent-89}, a ``natural'' dilation maps any subset of vertices
to the vertices that are neighbors of a vertex in that subset. The
adjunct erosion is then the set of all vertices whose neighborhood is
included in the initial set. Intuitively, one can guess that dealing
also with the edges of a graph can help for reaching a better
precision \citep{MA-07,ML-07,CNS-09,CNDS-13}. This was the motivation
for defining the analog ``natural'' dilation of a subset of edges
\citep{CNS-09,CNDS-13}: it contains all edges which are adjacent to
(\ie{} which share a common vertex with) an edge in the initial
subset. The adjunct erosion of a subset of edges contains each edge
whose neighborhood (\ie{} the set of all edges adjacent to a given
edge) is included in the initial subset. Interestingly, when one
applies simultaneously the vertex and edge natural dilations to the
vertex and edge sets of a subgraph, the resulting pair of edge and
vertex sets is still a subgraph, thus defining a natural dilation on
subgraphs \citep{CNS-09,CNDS-13}. The adjunct erosion is obtained by
the simultaneous applications of the vertex and edge erosions.

From a methodological viewpoint, in the usual framework of
mathematical morphology, one has to choose a structuring element that
parametrizes the operator. With morphology on graphs, the choice of a
structuring element is, in general, replaced by the choice of the edge
set that indicates which data are connected (see \citep{HNTV-92,HV-92}
for a framework of morphology on graphs where one must choose both an
edge set and a second ``graph'' that plays the role of a structuring
element). In the digital setting, there is a direct correspondence
between these two approaches. However, the use of graphs opens the
door to the processing of many kind of data (as seen in
\Xsec{sec:graphs}) and to new operators such as those described in the
next paragraphs.

The natural operators described above can be redefined and enriched
through the use of four elementary operators that are building blocks
(introduced in \citep{MA-07,ML-07} and further studied in
\citep{CNS-09,CNDS-13}) for morphology on graphs:
\begin{enumerate}
\item the vertex-edge dilation is a dilation that maps any set of
  vertices to the set of edges that contain at least one of these
  vertices;
\item the edge-vertex erosion, which is the adjunct erosion of the
  previous vertex-edge dilation, maps any set of edges to the set of
  vertices completely surrounded by edges of this set of edges (\ie{},
  vertices whose adjacent edges all belong to this set of edges);
\item the edge-vertex dilation is is a dilation that maps any set of
  edges to the set of vertices which are contained in one of these
  edges; and
\item the vertex-edge erosion, which is the adjunct erosion of the
  previous edge-vertex dilation, maps any set of vertices to the set
  of edges whose two extremities lie in the initial set of vertices.
\end{enumerate}
The natural dilation on vertices ({\em resp.}, edges) is simply the
composition of the vertex-edge ({\em resp.}, edge-vertex) dilation and
the edge-vertex ({\em resp.}, vertex-edge) dilation, whereas the
associated erosion on vertices ({\em resp.}, edges) is the composition
of the vertex-edge ({\em resp.}, edge-vertex) erosion and the
edge-vertex ({\em resp.}, vertex-edge) erosion. Since the four
operators defined above can be grouped as pair of adjunct operators,
they also lead to openings and closings. For instance, the successive
application of the vertex-edge dilation and the edge-vertex erosion is
the closing which, given a set of vertices, fills in the points which
do not belong to the set but which are completely surrounded by that
set (\ie{} the points whose (strict) neighborhood is completely
included in that set). Note that this closing is not the same as the
one obtained by composition of the natural dilation and erosion. In
fact, one can prove that the results of the two closings are ordered
(when applied to the same subset of vertices the result of the first
one is always included in the result of the second one). This leads to
interesting granulometries and alternating sequential filters.

The composition of any two dilations is still a dilation. Hence, by
successive applications of elementary dilations (a same dilation can
possibly be applied several times), one obtains series of dilations,
adjunct erosions, openings and closings. When the dilations used in
the compositions are those described in the previous paragraphs
(\ie{}, the natural dilations or the vertex-edge and edge-vertex
dilations), the associated series of closings ({\em resp.}, openings)
is ordered: when applied to a same object, the result obtained with
one closing ({\em resp.}, opening) of the series is always smaller
({\em resp.}, greater) than the result obtained with the next closings
of the series. These series of openings and closings, called
granulometries, are interesting for studying size distributions of
subsets of vertices, subsets of edges and subgraphs of a graph (see
\eg{} \citep{RS-10,CT-10}). Furthermore, from granulometries, series of
alternating sequential filters can be derived: each of them is a
sequence of intermixed openings and closings of increasing size. These
operators (which, contrarily to openings and closings, are not
extensive or anti-extensive) progressively filter the objects in a
balanced and progressive way. They constitute interesting tools for
simplifying subsets of vertices, subsets of edges and subgraphs of a
graph. \Xfig{fig:IllusASF} (top row) presents the result of such a
filtering procedure for a subset of pixels considered in the
4-adjacency graph. In this illustration, the edge-vertex and
vertex-edge dilations were used to obtain the alternating sequential
filters. As detailed in \citep{CNDS-13}, if, instead of the edge-vertex
and vertex-edge dilations, the natural dilations were used, then the
resulting filter would be less performing.

The morphological operators presented in the previous paragraphs are
all increasing. As such, they all induce stack operators acting on
functions weighting the vertices and/or edges of a graph (see
\citep{WCG-86} for stack operators,
\citep{Serra-82,MS-87,Heijmans-91,Ronse-06} for stack operators in the
context of flat mathematical morphology, and
\citep{Bertrand-05,Bertrand-07} for stack operators in the context of
watershed image segmentation). This allows for the definition of
morphological operators for weighted graphs, and thus for grayscale
images, to be systematically inferred from the ones on non-weighted
graphs (see \citep{CNDS-13}). The idea is to decompose a function into
level-sets by thresholding, then to apply a same operator to each
level-set, before reconstructing a resulting function by ``stacking''
these results. \Xfig{fig:IllusASF} (bottom row) presents the results
obtained with the grayscale extension of the graph alternating
sequential filters presented in the previous paragraph. Here the
operator is applied to a grayscale image structured by a spatially
variant graph obtained by removing the edges of the 4 adjacency graph
connecting two pixels with a high difference of intensity.

\section{Paths and shortest paths}
\label{sec:paths}

A classical problem in graph theory is to find a shortest path linking
two points~\citep{dijkstra1959} (Note that there may exist several such
shortest paths). It is not surprising that paths and shortest paths
find many applications in image processing and computer
vision~\citep{peyre-fnt-10}.

In a graph, a path is a sequence of vertices such that any two
successive vertices are linked by an edge. Depending of the
applicative context, several notions of length can be associated to
paths. The simplest one, when weights are not considered, consists of
counting the number of edges in the path. When weights are associated
to edges, one can for instance sum the edge weights along the path or
consider the maximum edge weights of the
path~\citep{pollack1960letter,udupa1996fuzzy}. Similar strategies can
be adapted for vertex-weighted graph. An optimal or shortest path
between two points is then a path of minimal length among all the
paths linking these two points. In graph theory, finding the length of
the shortest paths from a given vertex to all other vertices of the
graph is a well-studied problem. When the weights are always
positive\footnote{This condition can be relaxed, see
  \citep{falcao2004image}; see also the Bellman-Ford algorithm
  \citep{LRCC-01}.}, the algorithm proposed by \citet{dijkstra1959} provides an efficient solution.

An elementary use of paths is the computation of a distance map: from
any pixel of an image, one can compute the distance (length of the
shortest path) to the nearest obstacle vertex; labeling every vertex
with this distance provides what is called a {\em distance
  transform}~\citep{rosenfeld1968distance,fabbri20082d}. A common
obstacle vertex is a pixel of an object in a binary image. An
interesting property of distance maps is the following: a thresholding
of a distance map for a given value $m$ yields a dilation of size $m$
of the object. If the graph is unweighted, then the dilation is
exactly the natural dilation on vertices described in the previous
section~\citep{Vincent-89}. If the graph is weighted, then we still get
an algebraic dilation, however with a different geometric outcome.  In
general, a binary object dilated of size $m+n$ on a weighted graph is
not equal to the dilated of size $m$ of the same object dilated of
size $n$. As a special case, this composition law holds true for the
dilations on non-weighted graphs.

A notable use of paths is for morphological
filtering~\citep{heijmans2005path} of images that depict thin objects
of interest. Path openings and closings are algebraic morphological
operators using families of paths. Indeed, paths are thin and oriented
structuring elements that are not necessarily perfectly
straight. Hence, paths openings and closings offer more flexibility
than line-based openings and closings. Several variations around that
notion have been explored in the
literature~\citep{talbot2007efficient,cokelaer2012efficient,morard2014parsimonious}
and in
applications~\citep{valero2010advanced,tankyevych2010filtering,Morard2012}.

Many other usages of paths can be found in the literature. A
pioneering work~\citep{vincent1998minimal} aims at finding linear
features in images as optimal paths. A more recent and popular
contribution, called {\em seam carving}~\citep{avidan2007seam}, is
aimed at content-aware image resizing. A seam is an optimal path
connecting two image borders, either from top to bottom (vertical) or
from left to right (horizontal). The length of a seam is given by a
measure of contrast of the pixels along the path. Removing the least
important seams removes redundant part of the image, and thus makes it
possible to resize the image without distorting its content. Other
applications of the very same idea include contour
extraction~\citep{falcao1998user} (optimal paths between two seed
points are good contour candidates) and segmentation and
matting\footnote{Matting refers to the problem of accurate foreground
  estimation.}~\citep{saha2001relative,falcao2004image,bai2007geodesic}
(specifying several user-provided seeds, each region of the
segmentation is given by the vertices that are closest to one of the
seeds with respect to all the others seeds).

\begin{figure*}
\centering
\begin{tabular}{c}
\begin{tabular}{ccc}
\subfigure[]{\includegraphics[width=0.25\linewidth]{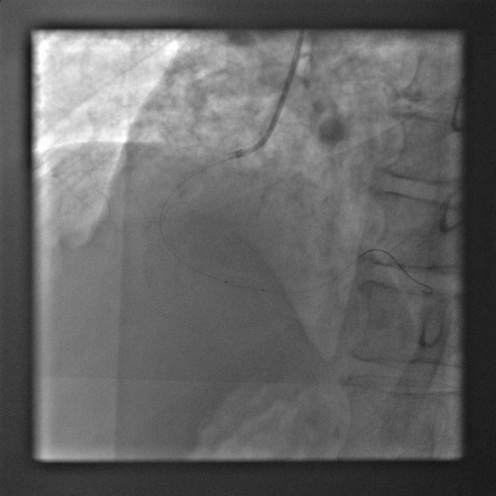}}
\subfigure[]{\includegraphics[width=0.25\linewidth]{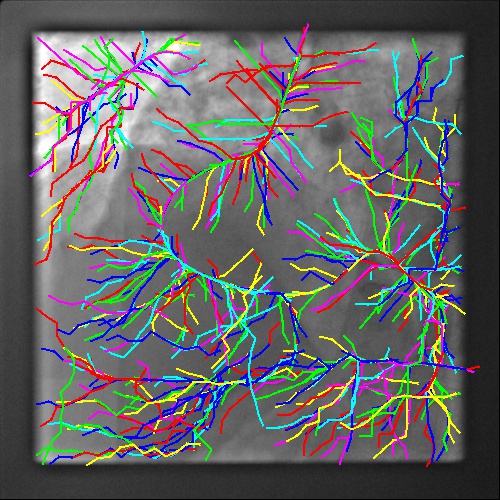}}
\subfigure[]{\includegraphics[width=0.25\linewidth]{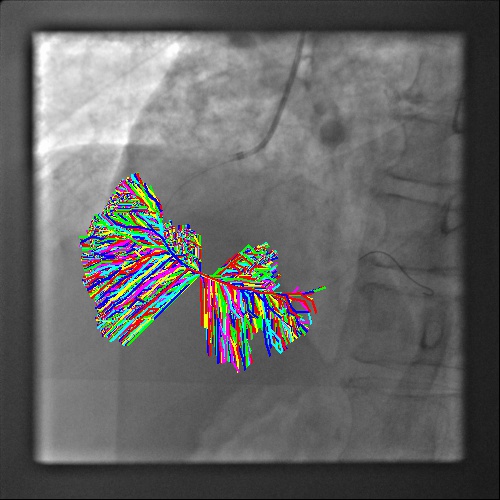}}
\end{tabular}\\
\begin{tabular}{cc}
\subfigure[]{\includegraphics[width=0.25\linewidth]{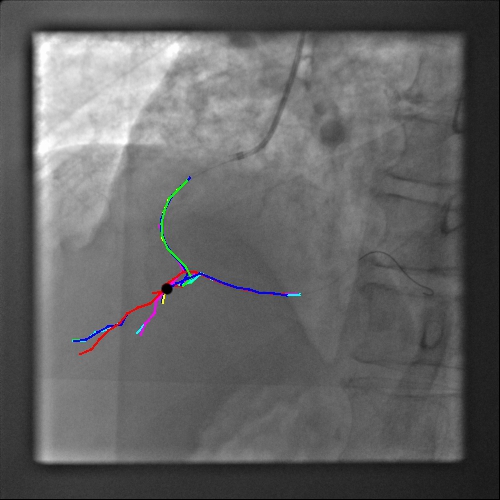}}
\subfigure[]{\includegraphics[width=0.25\linewidth]{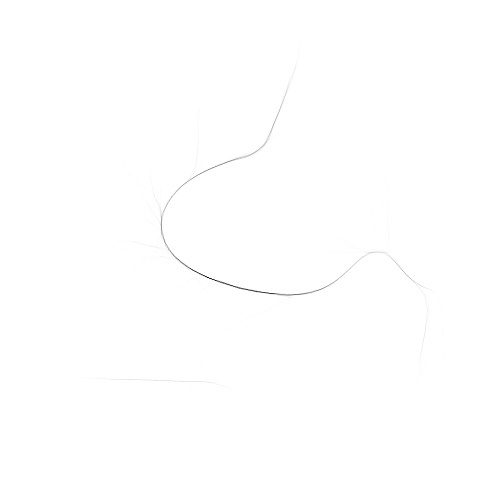}}
\end{tabular}
\end{tabular}
\caption[ ]{(a) X-ray fluoroscopy image from an angioplasty exam
  illustrating a guide-wire, with a long smooth curve appearance and
  low contrast to noise ratio~\citep{bismuth2012curvilinear}. (b): 500
  locally optimal paths originating from random locations. Observe
  their tendency to converge to the linear structures of the image and
  especially to the guide-wire. (c,d) The set of paths intersecting at
  one given point (belonging to the guide-wire, in (c), and to the
  background in (d), in this case the point is indicated by the dark
  spot). (e) The result of path voting perfectly finds the elongated
  structure. See~\citep{bismuth2012curvilinear,bismuth:tel-00786326}
  for more details.}
\label{fig:PPI}
\end{figure*}

More generally, one can compute for any pixel of the image, an optimal
path of a given length. By selecting several seeds, one obtain an
image of paths that has many
applications~\citep{cohen1997global}. Choosing the correct seed pixels
is in general application-dependent~\citep{rouchdy2008image}.  An
interesting choice is to choose as seeds all image
pixels~\citep{bismuth2012curvilinear}. We can also add some regularity
constraints on the paths, for example, we can request them to be
polygonal: indeed, polygonal paths are less tortuous than usual
optimal paths. Polygonal Path Images~\citep{bismuth2012curvilinear}
(PPI) are useful tools for enhancing thin objects in images: for
example, one can count the number of paths of the PPI that run through
a given pixel; the higher this number, the higher the probability of
presence of an actual thin object (see Fig.~\ref{fig:PPI} for an
example). Other uses of such maps are described in
\citep{bismuth:tel-00786326}.

As seen in this section, a great variety of powerful image operators
can be implemented using optimal paths In the context of graph-based
image processing applications, this approach has been promoted notably
under the name of image foresting
transform~\citep{falcao2004image,falcao2004interactive,PFAT-12}
(IFT). In particular, IFT allows the implementation of operators based
on connectivity: region growing, ordered propagation, watershed,
flooding, geodesic dilation, morphological reconstruction, etc. The
IFT framework is thus a first unifying framework for presenting such
operators. In the next section, we detail another framework for
connected operators, based on optimum spanning forests.


\section{Connected filters, watersheds and hierarchies}
\label{sec:conwat}

In this section, we review morphological segmentation
(\Xsec{sec:segwat}) and filtering methods (\Xsec{sec:confilt}) that
rely on the notion of connected components. These segmentation and
filtering methods are deeply related: in general, the filtering
methods lead to interesting segmentation in (quasi) flat zones whereas
the segmentation methods lead to a cartoon (filtered) image where all
vertices of a region take a constant value such as the mean of the
original values in the region. In many cases, when the results depend
on a scale parameter, the set of all possible results are organized as
a hierarchy (\Xsec{sec:hiermst}). We conclude the section with an
important practical point, the design of criteria adapted to the
task (\Xsec{sec:criterion}).

\subsection{Segmentation: flat zones, watersheds and minimum spanning
forest} 
\label{sec:segwat}
Image segmentation is the task of delineating objects of interest that
appear in an image, or more generally in a graph. In many cases, the
result of such a process, also called a segmentation, is a set of
connected regions which are composed of vertices, and are separated by
a frontier. Depending on the applicative context, the frontier set can
be made of vertices or can be an inter-vertices separation made of
edges. In the first case, a formal notion of frontier is the one given
by a binary watershed or cleft~\citep{Bertrand-05, CBCN-08} and in the
second case graph cuts~\citep{Diestel-97,BVZ-01,CBNC-09} are considered
as frontiers. In all cases, a region or a set of vertices is connected if
there exists a path that is included in the region and that links any
two of its vertices. A connected set is furthermore a connected
component of the graph if none of its proper supersets is still
connected. The notion of a connected component in a graph is
fundamental for defining two basic morphological segmentation methods:
the quasi-flat zones and the watersheds.

The flat zones segmentation partitions the vertices of a non-negative
edge-weighted graph. The partition is obtained as the set of connected
components of the graph whose vertices are those of the weighted graph
and whose edges are those with a null weight in the weighted graph.
When the weight function is the gradient (see \Xsec{sec:graphs}) of a
grayscale image, the gray level in each flat zone is constant, and the
flat zones are the maximal connected sets satisfying this property. In
many cases, the flat zones segmentation is too fine (\ie{}, contains
too many small regions) and quasi-flat zones may be better adapted
(see \eg{} \citep{NMI-79,MM-99,soille2008pami}). To this end, the
connected components are considered in the graph whose edges are those
with weight below a given positive value. As we will see later in this
section, (quasi-) flat zones are the basis for powerful hierarchical
segmentation and filtering methods.

\begin{figure*}[htb]
    \begin{center}
    \begin{tabular*}{1\linewidth}{@{\extracolsep{\fill}}c c}
      \includegraphics[width=0.4\linewidth]{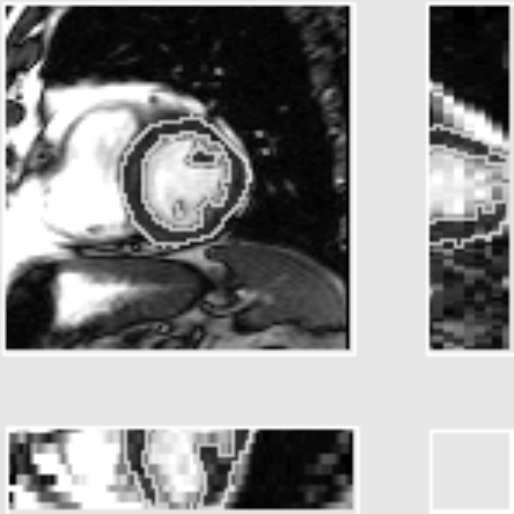}&
      \includegraphics[height=0.4\linewidth]{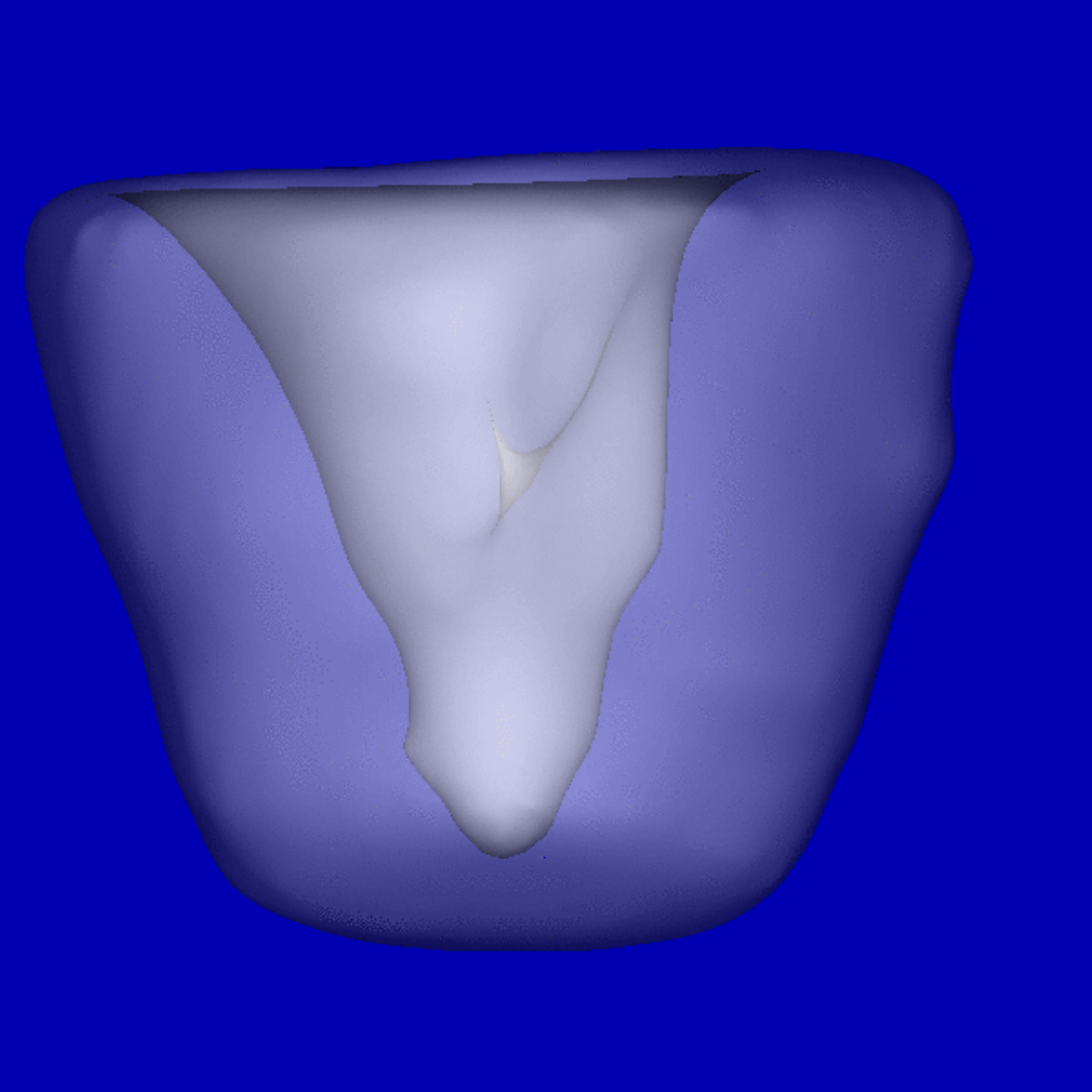}\\
\end{tabular*}
\caption{\label{fig:lpeCoeur} Example of 3D+t (time+space) left
  ventricular myocardium segmentation by watershed cuts. Left: three
  orthogonal sections of a cardiac 3D MRI superimposed with the
  internal border of the segmented left ventricular myocardium. Right:
  a three dimensional rendering of the segmented object. The watershed
  cut is computed in 4D (considering time as a supplementary dimension
  to the space) from markers obtained by a series of morphological
  operators and the resulting regions are smoothed by alternating
  sequential filters (see more details in \citep{CNCCGG-10}). This
  method has been validated by comparisons with manual segmentation
  performed by cardiologists \citep{CNCCGG-10} and by comparisons with
  other state-of-the-art methods~\citep{IMPEIC-12}.}
  \end{center}
\end{figure*}

\begin{figure*}[htb]
  \begin{center}
   \begin{tabular*}{1\linewidth}{@{\extracolsep{\fill}}c c c}     
      \includegraphics[width=0.3\linewidth]{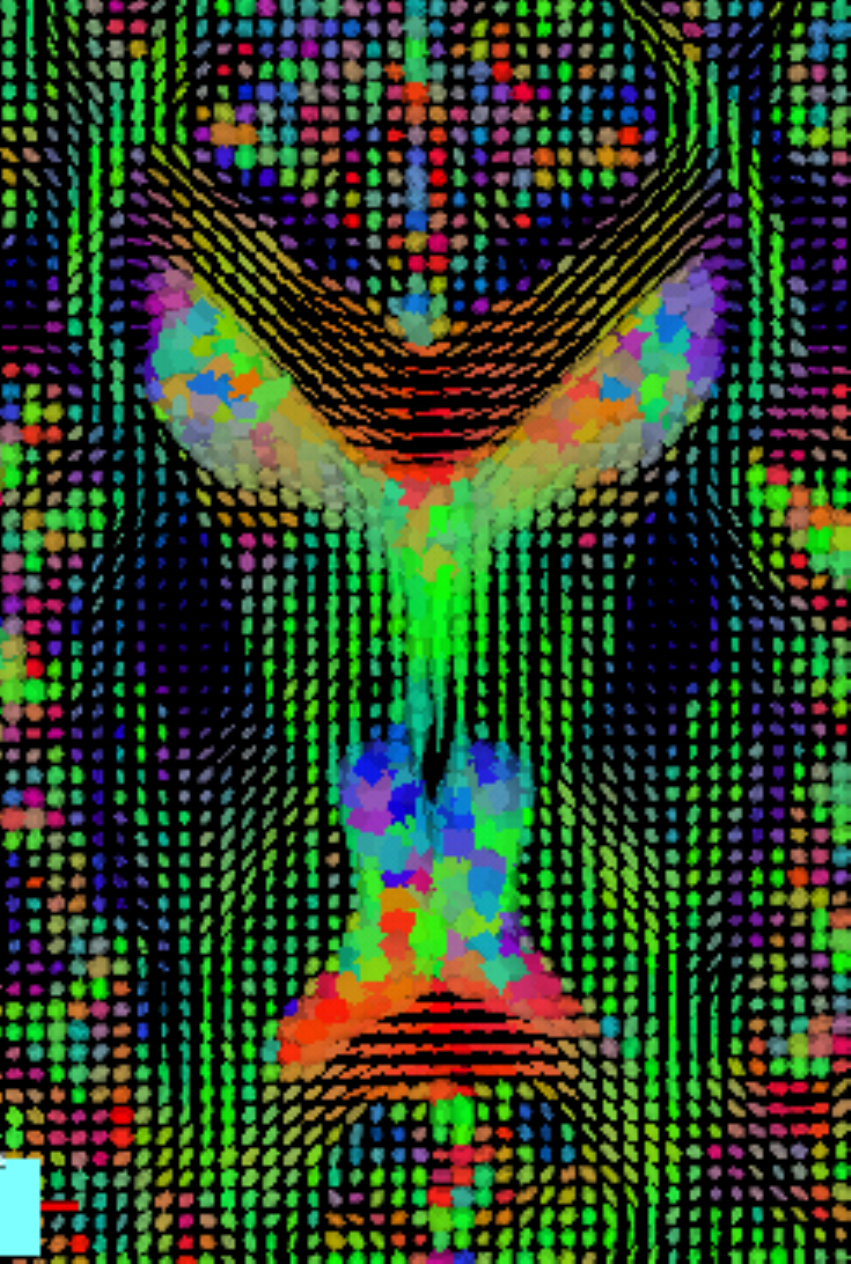}&
      \includegraphics[width=0.3\linewidth]{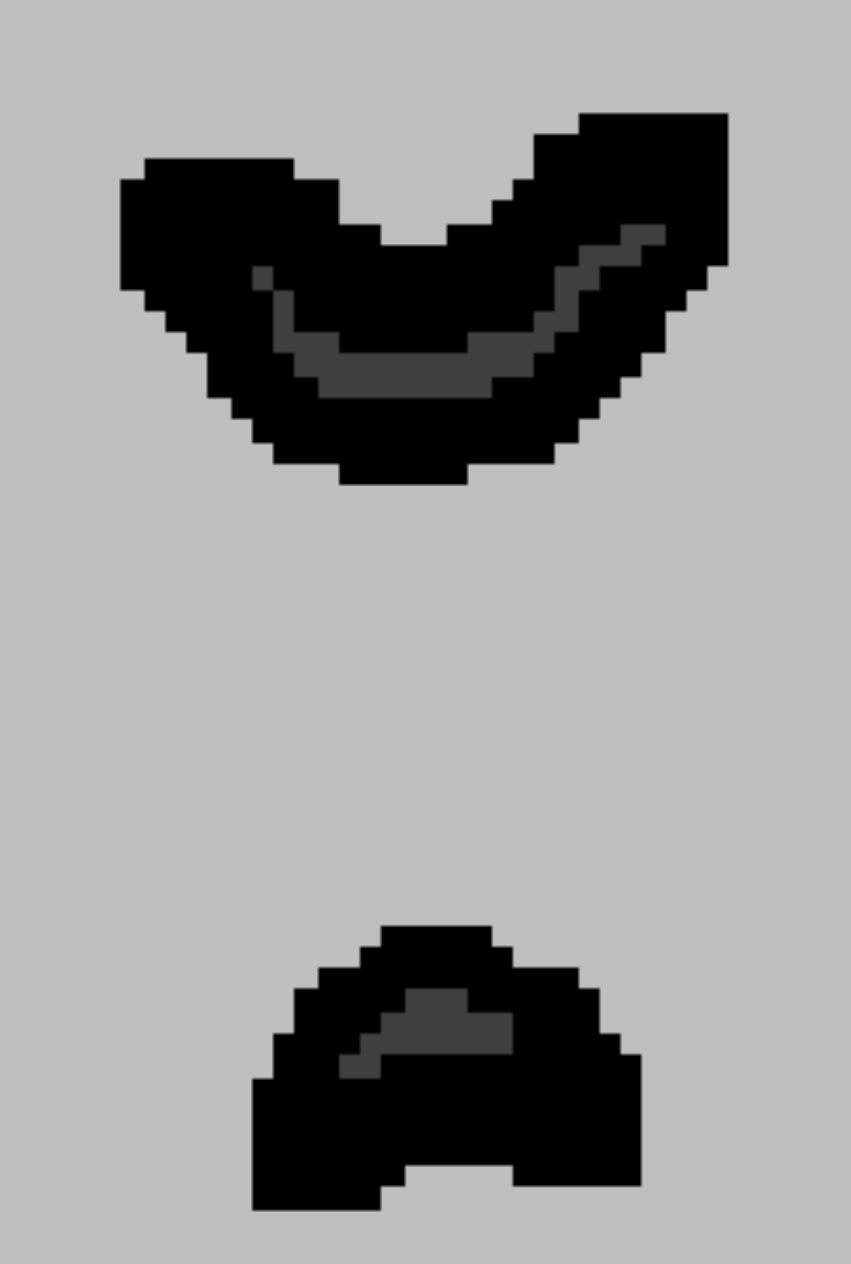}&
      \includegraphics[width=0.3\linewidth]{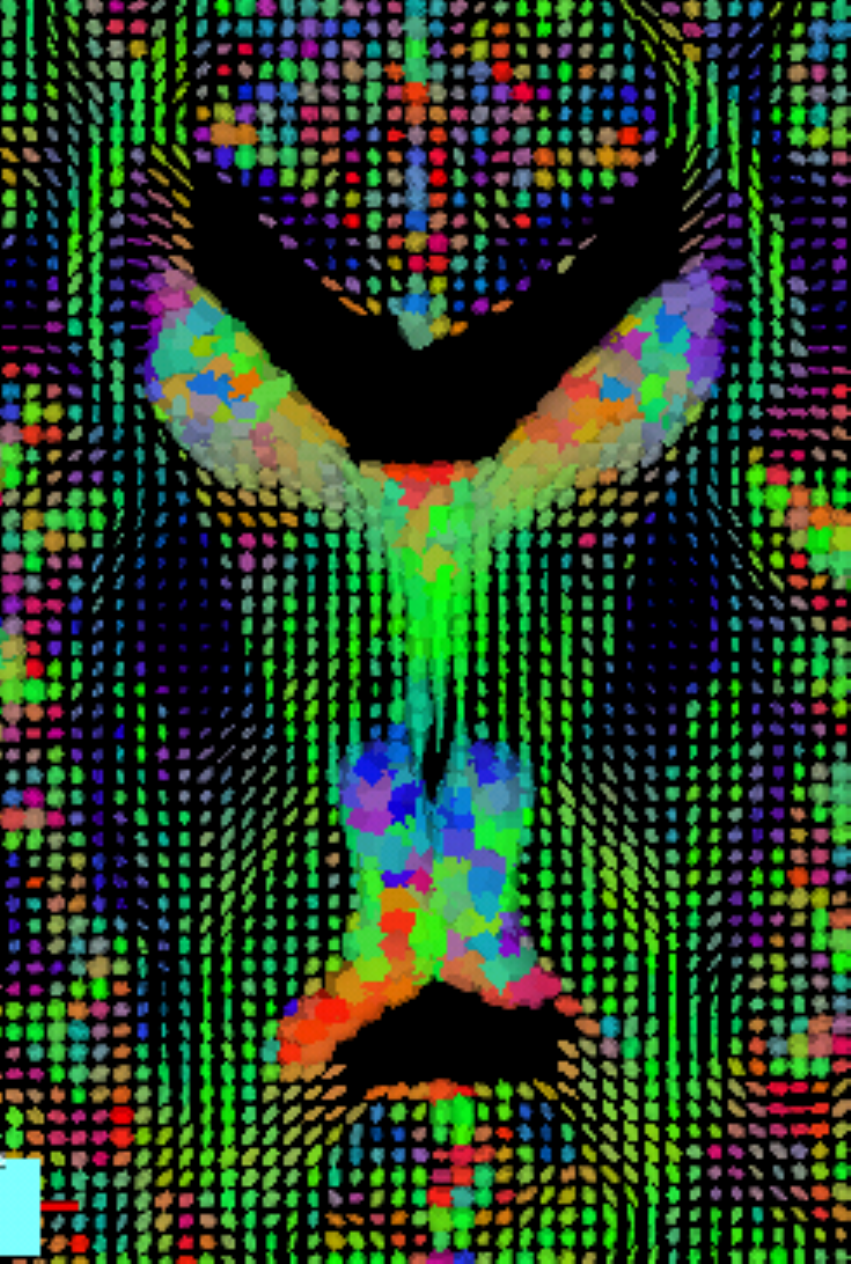}\\
      $(a)$ & $(b)$ & $(c)$
    \end{tabular*}
    \caption{\label{fig:tensorSeg} Illustration of Diffusion Tensor
      Images (DTIs) segmentation. $(a)$: A close-up on a cross-section
      of a 3D brain DTI. $(b)$: Image representation (in the same
      cross-section as~$(a)$) of the markers, obtained from a
      statistical atlas, for the corpus callosum (in dark gray) and
      for its background (in light gray) $(c)$: Segmentation of the
      corpus callosum by a marker based watershed cut. The tensors
      belonging to the region corresponding to the seed labeled
      ``corpus callosum'' are removed from the initial DTI and thus
      the corresponding voxels appear black (see more details about
      this illustration in \citep{CBNC-10} and about DTI morphological
      segmentation in \citep{RL-08}). }
  \end{center}
\end{figure*}

The watershed transform introduced by \citet{BL-79} for morphological
segmentation and later popularized by \citet{VS-91} is used as a
fundamental step in many image segmentation procedures. A grayscale
image, or more generally a function, is seen as a topographic surface:
the gray values become the elevations, the basins and valleys
correspond to dark areas whereas the mountains and crest lines
correspond to light areas. Intuitively, the watershed is a subset of
the domain, located on the ridges of the topographic surface, that
delineates its catchment basins. It may be thought of as a separating
line-set from which a drop of water can flow down towards several
minima. For applications to image segmentation, the watershed is often
computed from the gradient magnitude of an image. Therefore, the
resulting contours are located on high gradient contours of the image,
which often correspond to the borders of the objects of interest (see
\eg{}, \Xfigs{fig:lpeCoeur} and~\ref{fig:tensorSeg}).

Following the intuitive drop of water principle presented in the
previous paragraph, the watershed cuts, a notion of a watershed in
edge-weighted graphs, were introduced in \citep{CBNC-09}. A watershed
cut is indeed a graph cut: it is only made of edges and it partitions
the vertex set of the underlying graph. The consistency of watershed
cuts was established by~\citet{CBNC-09}: they can be equivalently
characterized by their catchment basins (through a steepest descent
property) or by their dividing lines (through the drop of water
principle). In a discrete framework, watershed cuts are the first
watershed definition that satisfies this natural consistency
property. Furthermore, a global optimality property of watershed cuts
is provided in \citep{CBNC-09} by an equivalent characterization in
terms of minimum spanning forests.

The minimum spanning tree (MST) problem \citep{LRCC-01} is one of the
most typical and well known problems of combinatorial optimization:
given a connected edge-weighted graph, find a connected subgraph that
is spanning (\ie{} whose vertex set is the same as the given edge
weighted graph) and whose weight is minimal, the weight of a subgraph
being the sum of the weights of its edges. A minimum spanning tree of
an edge-weighted graph can be computed by efficient and easy to
implement algorithms \citep{Boruvka-26,Kruskal-56,Prim-57}. For
tackling image segmentation problems, we are interested by optimal
structures that are not necessarily connected since we look for
segmentations made of several connected regions. In this case, minimum
spanning forests (MSF) are adapted: given a set of ``root'' vertices,
a MSF is a minimum weight subgraph among the family of all spanning
subgraphs such that each connected component contains exactly one
root. The first links between watershed segmentations and MSFs were
drawn by \citet{Meyer-94}. Later \citet{CBNC-09} proved that the
catchment basins provided by watershed cuts and the connected
components of the MSFs rooted in the regional minima of the weight map
are the same. As we will see in the next section, minimum spanning
trees are also deeply related to hierarchical segmentations or more
generally to hierarchical representations of data.

Additionally, watershed cuts have also been characterized in terms of
shortest paths \citep{CBNC-10}, drawing a link with the IFT framework
described in the previous section (see also
\citep{audigier2007watershed} for a link between minimum spanning
forest and IFT) and in terms of flooding \citep{CBNC-10}, making a link
with the watershed presentation popularized in the 90's
\citep{VS-91,meyer-beucher90,Meyer-91,BM-92}. Links between watershed
cuts and other popular graph based segmentation methods such as
min-cuts or random walks were established in \citep{AACK-10} and
\citep{CGNT-11} respectively. As far as we know, similar properties
have not been obtained in other discrete settings. In particular, when
one wants to obtain as a watershed a separation made of vertices, this
results in weaker properties (see counter examples in
\citep{NCB-05,Cousty-07}). Among the watershed definitions or
algorithms producing a separation made of vertices, the topological
watershed \citep{CB-97}, which is defined for vertex-weighted graphs,
can be characterized by interesting properties of contrast
preservation \citep{Bertrand-05,Bertrand-07}. It was shown
in~\citep{CBNC-10} that these properties are also satisfied by
watershed cuts. Given a weight function, it must be noted that there
exist, in general, several watersheds. The choice of one of these
watersheds can be arbitrary or based on a (optimal) criterion (see
discussions related to this subject in
\citep{MN-10,AL-07,CGNT-11,SPKH-13}).

Efficient algorithms for computing watersheds is an intense subject of
research since its introduction in the late 70's.  \citet{VS-91},
followed by \citet{Meyer-91}, were the first to propose linear-time
complexity watershed algorithms relying on sorting the pixels
according to their gray level and on a hierarchical priority queue,
respectively. The topological watershed \citep{CNB-05} can be computed
in quasi-linear time thanks to the min tree (see \Xsec{sec:confilt})
of the function. The watershed cuts can be obtained in linear time
\citep{CBNC-09,CBNC-10}, without any sorting or auxiliary data
structures such as a hierarchical queue or a component tree. The
interesting trade-off between the precision of the watershed contours
and the low computational costs is an important reason for the
popularity of watersheds in applications.

When the methods described in this section are applied for analyzing
an image, they often produce an over-segmentation: the obtained
partitions are too fine and contain more regions than objects of
interest appearing in the image. Marker-based (or seed-based)
segmentation is a usual procedure to prevent this
over-segmentation. Given a set of ``seed'' or ``root'' vertices, which
mark regions of interest in the image, the idea is to obtain a cut or
a partition of the vertices such that each region contains exactly one
seed. In mathematical morphology, this methodology is presented and
developed in \citep{meyer-beucher90,BM-92} under the name of watershed
from markers. Given a set of seeds, one can modify an image or
function so that after this filtering, the segmentation of the
transformed function is a partition associating exactly one region to
each seed. For instance, in a seeded watershed procedure, one needs a
function such that regional minima correspond to seeds. Connected
filters, which are described in the next section, allow this kind of
filtering to be performed. They also allow for producing functions
such that the associated segmentations are made of exactly~$k$
regions, where~$k$ is a predefined value. The obtained regions are
then the most significant according to a certain criterion used for
the filtering step.

\subsection{Connected filtering} \label{sec:confilt}


In binary morphology, connected filters act by removing specific
connected components of a graph, while leaving the remaining connected
components perfectly preserved. The extension to weighted graphs is
straightforward when we consider stacks, as described in
section~\ref{sec:adjunc}. For example, if we want to remove all round
white objects from the graph, we first design an attribute or a
(numerical) criterion that states how round is a component; then we
consider the family ${\cal C}$ of all the connected components of all
the upper level sets of the weighted graph, and we remove the
components that are not round enough for the criterion. We can then
reconstruct a filtered weighted graph with the remaining
components. From an algorithmic standpoint, an efficient
implementation relies on the fact that the family ${\cal C}$
can be structured in a tree, called the max-tree in the
literature \citep{PS-salembier_IEEEIP98}.  Indeed, any
two connected components of ${\cal C}$ are either disjoint or
nested. There exists fast algorithms for computing this
max-tree~\citep{NC06,berger2007effective,Wilkinson11}, see
\citep{carlinet2013comparison} for a survey and a comparison.

\begin{figure*}[htb]
 \centerline{
   \begin{tabular}{cccc}
     \includegraphics[width=1.8cm]{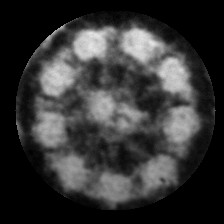}
   &
     \includegraphics[width=1.8cm]{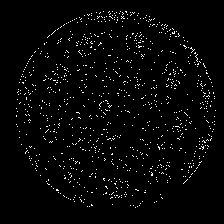}
   &
     \includegraphics[width=1.8cm]{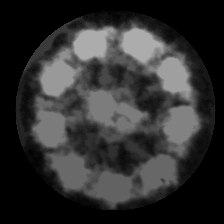}
   &
     \includegraphics[width=1.8cm]{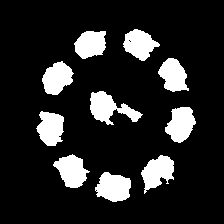}
   \\
   (a) & (b) & (c) & (d)
  \end{tabular}
 }
 \caption{(a) Original image. (b) Maxima of image (a), in white. (c)
   Image filtered with an increasing criterion (volume) on the
   max-tree. (d) Maxima of image (c), which correspond to the ten most
   significant lobes of the image~(a).  }
\label{fig:cellule}
\end{figure*}

\begin{figure*}[htbp]
  \begin{center}
    \begin{minipage}[!t]{0.60\linewidth}
      \begin{center}
        \centerline{\includegraphics[width=1.0\linewidth]{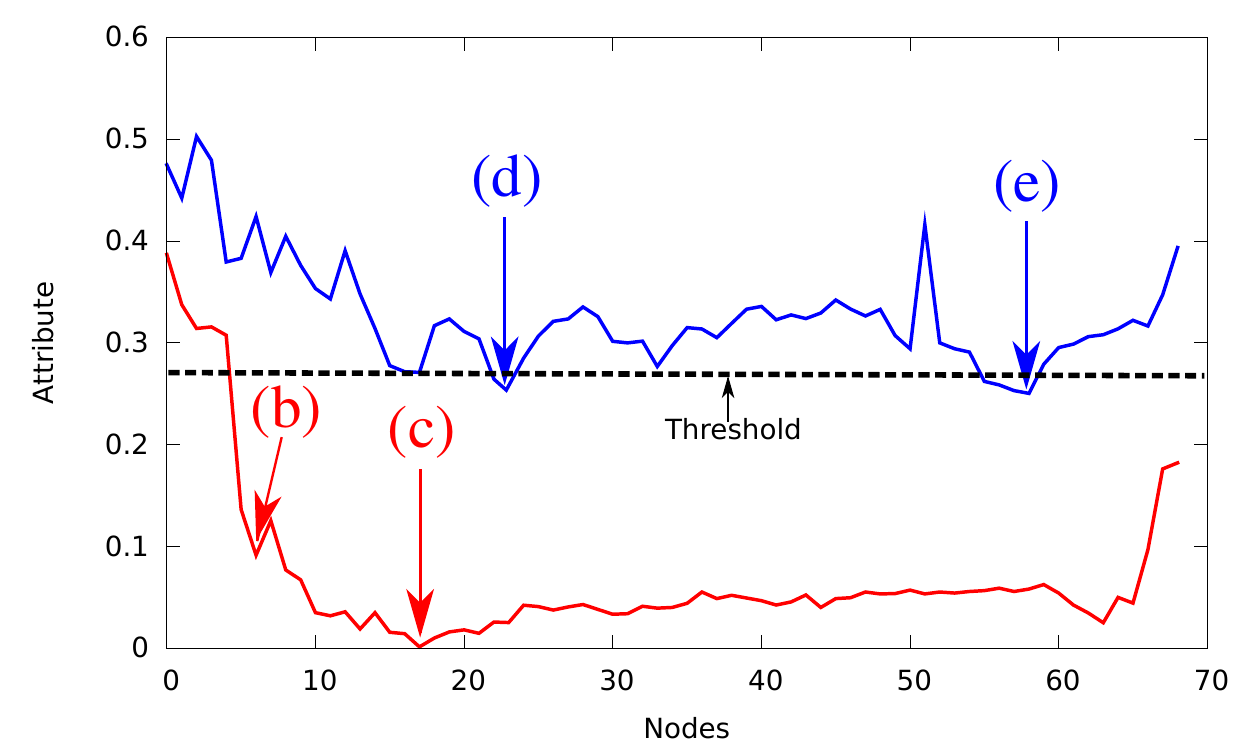}}
        \centerline{(a)}
      \end{center}
    \end{minipage}
    \begin{minipage}[!t]{0.36\linewidth}
      \begin{tabular}{ p{0.5\textwidth} p{0.5\textwidth} }
        \centerline{\includegraphics[width=0.9\linewidth]{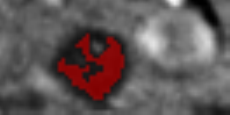}}
        &
        \centerline{\includegraphics[width=0.9\linewidth]{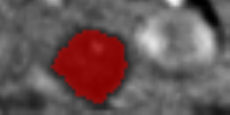}}
        \tabularnewline
        \centerline{(b)} & \centerline{(c)}
        \tabularnewline

        \centerline{\includegraphics[width=0.9\linewidth]{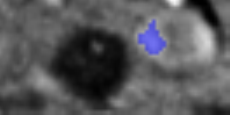}}
        &
        \centerline{\includegraphics[width=0.9\linewidth]{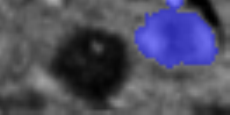}}
        \tabularnewline
        \centerline{(d)} & \centerline{(e)}
        \tabularnewline

        \centerline{\includegraphics[width=0.9\linewidth]{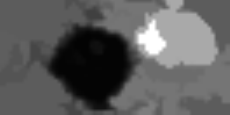}}
        &
        \centerline{\includegraphics[width=0.9\linewidth]{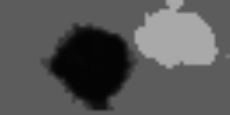}}
        \tabularnewline
        \centerline{(f)} & \centerline{(g)}
      \end{tabular}
    \end{minipage}
    \caption{(a) Evolution of a ``circularity'' criterion on two
      branches of a tree of shapes~\citep{xu2013two}; (b to e): Some
      shapes; (f) Attribute thresholding; (g)~A~morphological
      shaping.}
    \label{fig:circularity}
  \end{center}
\end{figure*}

From a high-level standpoint, such a filtering is equivalent to a
thresholding of the max-tree, seen as a node-weighted graph whose
nodes are the components and weights are given thanks to the
criterion. When the criterion is increasing (meaning that if a
connected component $A$ is included in another component $B$, then its
attribute is lower than the attribute of $B$), the thresholding
amounts to cutting branches in the tree (see Fig~\ref{fig:cellule} for
an example). However, the majority of useful criteria are not
increasing. Thresholding then removes nodes within a branch, and thus,
as classical image thresholding that does not take the pixel context
into account, is not very robust to noise: although two nodes in two
different branches of the tree can appear visually very similar, the
criterion can identify them as being very different (see
Fig.~\ref{fig:circularity}). Several strategies have been proposed to
robustify
filters~\citep{PS-salembier_IEEEIP98,urbach2007connected,salembier2009connected,Salembier2010},
they all amount to cutting whole branches of the tree: if a specific
node has to be removed, then all its descendant are also removed. A
fruitful and seminal idea, called {\em
  shaping}~\citep{XGN-12,xu2013two}, is to apply a connected filter on
the tree itself, seen as a weighted graph whose neighborhood
relationship is given by the parenthood relationship: a node is
neighbor of its parent and its children, and the weight is given
thanks to the criterion. We can then build a max-tree on this graph,
and use an increasing criterion on this second tree to robustly remove
components.

Other trees are possible, for example the min-tree, which is made from
all the connected components of the lower-level sets. The min-tree
helps dealing with dark components. Both the max-tree and the min-tree
are also known as the component
tree~\citep{jones99,breen-jones96a,NC06}. Another tree example is the
so-called {\em tree of
  shapes}~\citep{monasse2000fast,caselles2010tree,Geraud2013,najman2013discrete},
which is intuitively the tree of all the level lines of a graph. The
tree of shapes deals with both white and black components at the same
time, and thus is useful in producing self-dual filters.  There are
numerous topological issues at play here, and this line of work is
intimately linked to what is done in (discrete) Morse
theory~\citep{forman2002user}, algebraic topology and persistent
homology~\citep{edelsbrunner2000topological} (see also
section~\ref{sec:beyond}).

\subsection{Hierarchies of partitions and optimum spanning forests} 
\label{sec:hiermst}



\begin{figure*}
  \begin{center}
    \begin{tabular*}{1\linewidth}{@{\extracolsep{\fill}}c c}
      \subfigure[]{
        \includegraphics[width=0.4\linewidth]{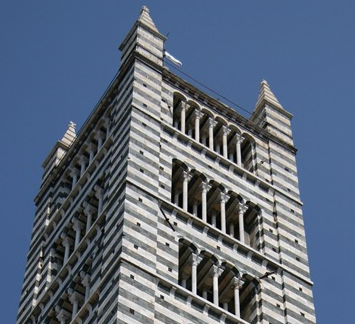}}&
      \subfigure[]{
        \includegraphics[width=0.5\linewidth]{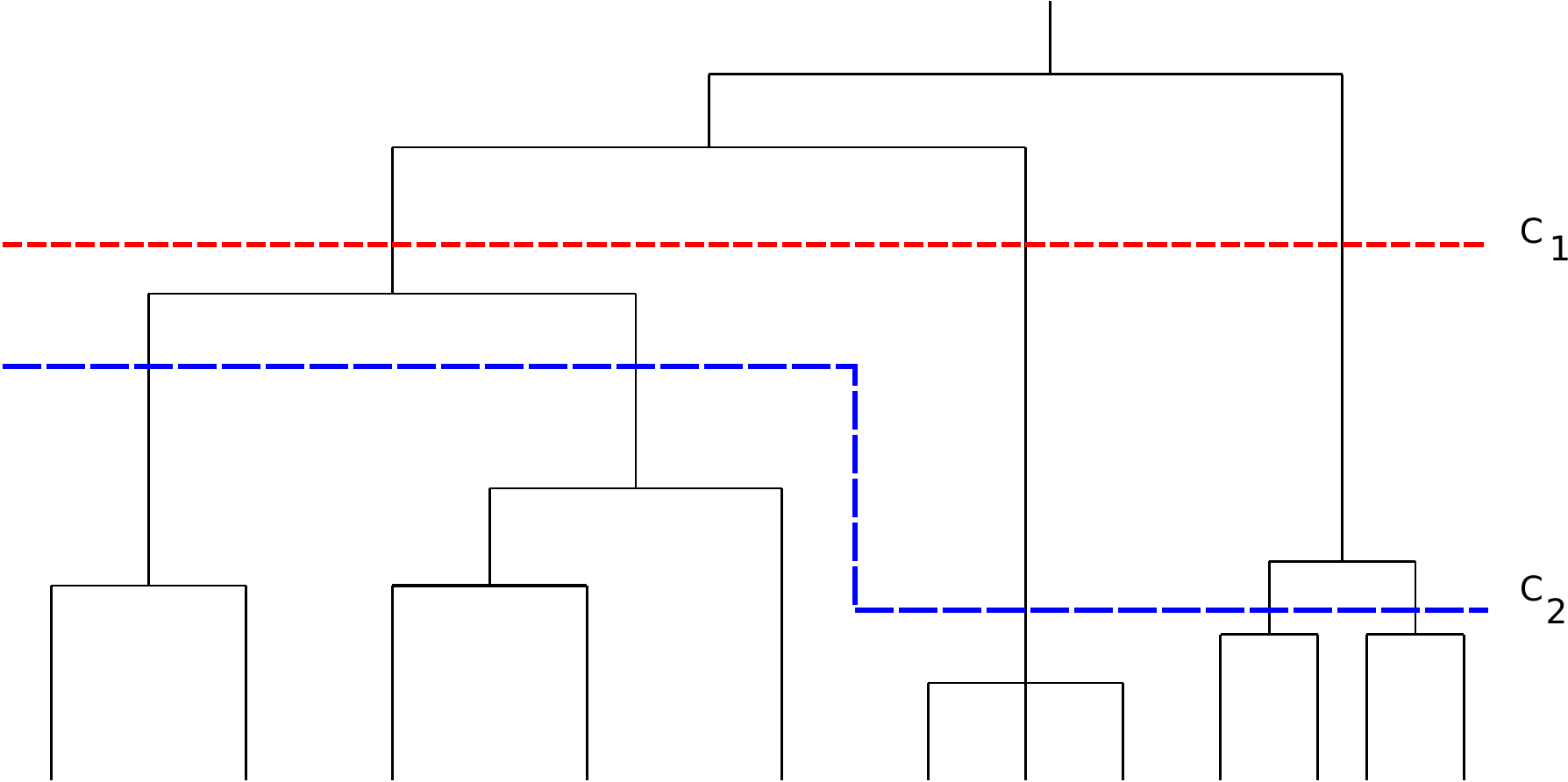}}\\
      \subfigure[]{
        \includegraphics[width=0.4\linewidth]{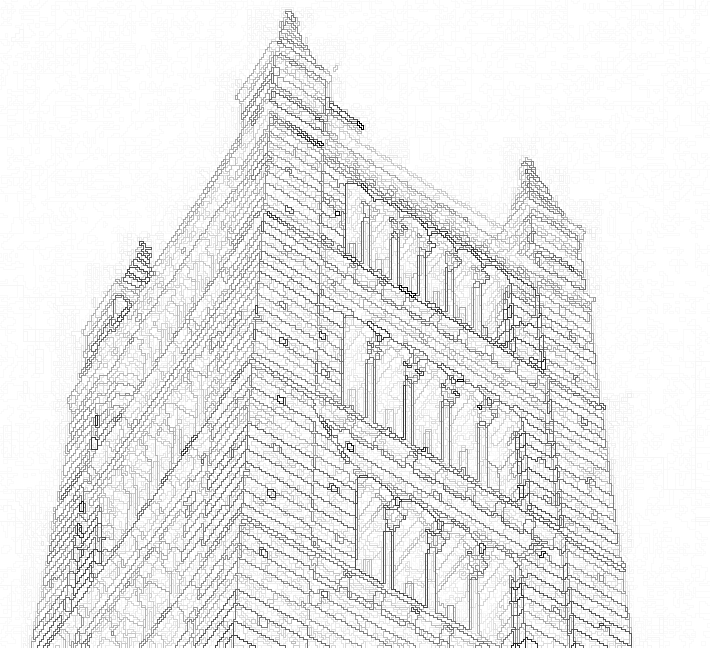}}&
      \subfigure[]{
        \includegraphics[width=0.4\linewidth]{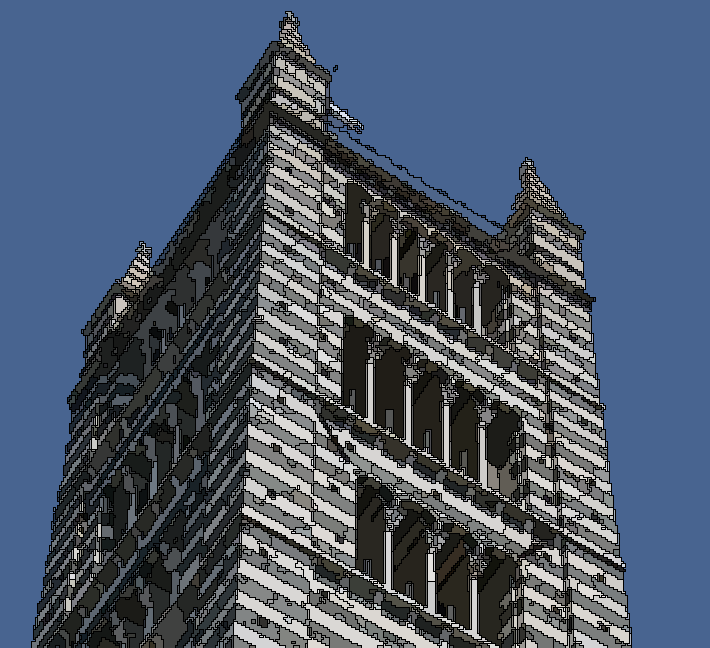}}
    \end{tabular*}
  \end{center}
  \caption{Hierarchical segmentation and filtering. (a) A color
    image. (b) A hierarchy of flat zones~\citep{SS95} of (a),
    represented by its dendrogram (min-tree of the minimum spanning
    tree of a color
    distance~\citep{cousty2013constructive,cousty2011incremental,najman2013playing}). The
    two cuts $C_1$ and $C_2$ correspond to two different flat-zone
    segmentations of (a), $C_1$ being a horizontal cut and $C_2$ being
    a non-horizontal cut~\citep{GuiguesCM06} (called a flooding in the
    morphological literature~\citep{MN-10}). (c) A saliency map
    \citep{NS96}, theoretically equivalent to the
    dendrogram~\citep{najman2011equivalence}, but with better
    visualisation properties. (d) A segmentation of (a) in which each
    region has been colorized by the mean color of the pixels forming
    the region. Such a coloring is a filtering of (a). The
    segmentation~(d) is obtained equivalently by either a thresholding
    of~(c) or by a horizontal cut of~(b). Other saliency maps can be
    obtained through floodings of~(d) or, equivalently, trough
    non-horizontal cuts of~(b). }
  \label{fig:hierarcuts}
\end{figure*}

In the previous section~\ref{sec:confilt}, we did not pay strict
attention to the type of graph under scrutiny. Indeed, the ideas can
be applied to any weighted graph, whether it is a vertex-weighted
graph or an edge-weighted graph. However, traditionally, connected
filters have been applied to vertex-weighted graph. But the very same
ideas can be applied to edge-weighted graphs. This has been a common
practice for at least 20 years, without always a clear realization
that this was indeed done. The main example has been mentioned before:
the quasi-flat zones hierarchy~\citep{SS95}. Any hierarchy can be
represented as a tree, called a dendrogram (see
Fig.~\ref{fig:hierarcuts}.b). In fact several trees can be used to
represent a given
hierarchy~\citep{cousty2013constructive,najman2013playing}. As
described in section~\ref{sec:segwat}, the quasi-flat zones hierarchy
is obtained by thresholding an edge-weighted graph, the weights being
a gradient of intensity. Two connected components of two threshold
levels are either disjoint or nested, hence the tree structure. It has
been shown in \citep{cousty2013constructive} that the connected
components of all the thresholds (organized with the inclusion
relationship) can be obtained from the min-tree of the edge-weighted
graph, which can be computed by efficient algorithms. But components
with exactly the same vertices can be obtained by considering only a
minimum spanning tree of the edge-weighted
graph~\citep{cousty2013constructive}, which uses less memory than the
original graph and is easier to handle because it contains less
redundancy~\citep{najman2013playing}. The min-tree of the minimum
spanning tree is called the alpha-tree in the literature, and
specific algorithms for computing it can be
designed~\citep{najman2013playing,havel2013efficient}.

Filtering the min-tree with an increasing criterion is a process that
is known as a {\em flooding} in the watershed
literature~\citep{meyer-beucher90,MN-10,cousty2008raising}. The very
same process has been done on the alpha-tree in the constrained
connectivity framework~\citep{soille2008pami} in the literature
(albeit without the link to the min-tree of the MST we just
mentionned). The reason to restrict ourselves to increasing criteria
is for transforming a hierarchy into another hierarchy. Indeed,
filtering a hierarchy amounts to do a {\em non-horizontal
  cut}~\citep{GuiguesCM06,MN-10} in the hierarchy (see
Fig.~\ref{fig:hierarcuts}.b). If a criterion (depending on a
parameter) is increasing, all the possible non-horizontal cuts (for
all the possible values of the parameter) stack, hence providing a
novel hierarchy. 

Hierarchies have been exploited in image processing and computer
vision since the beginning \citep{Zahn71,Morris1986,Pavlidis}.
However, many criteria used in practice are not increasing. A current
popular example of a non-increasing criterion is proposed
in~\citep{FH-04}; the criterion is based on measuring the
dissimilarity between elements along the boundary of the two
components relative to a measure of the dissimilarity among
neighboring elements within each of the two components. The algorithm
proposed in \citep{FH-04} extracts from the hierarchy of quasi-flat
zones a segmentation that is neither too coarse nor too fine. Several
attempts to produce a hierarchy based on the same criterion can be
found in the literature
\citep{haxhimusa2004segmentation,guimaraes2012hierarchical,xu2013two}. The
idea of doing a shaping, {\em i.e.} a connected filter on the
dendrogram of the hierarchy, seen as an node-weighted graph whose
weight is given by the criterion of \citep{FH-04}, is explored in
\citep{xu2013two}. A different approach is proposed in
\citep{guimaraes2012hierarchical}: the idea is to relax one of the two
constraints, for example one can extract, from the hierarchy of quasi
flat-zones, a (largest) hierarchy of segmentations that are not too
coarse, but these segmentations can be too fine.

To conclude this section, let us mention an interesting representation
of hierarchy of segmentations: it consists in stacking all the
contours of the segmentations, or equivalently, in valuating each
contour by the number of times it appears in the hierarchy (see
Fig.~\ref{fig:hierarcuts}.c and Fig~\ref{fig:statues}). This notion
has been introduced under the name of {\em geodesic saliency} of
watershed contours in \citep{NS96}, has been independently
rediscovered by \citet{GuiguesCM06}, and is extensively used in
\citep{arbelaez2011contour}, where it is the main entry point for
evaluating the quality of a given hierarchy. It has been proved
in~\citep{najman2011equivalence} that such a representation is
formally equivalent both to a specific watershed (called {\em
  ultrametric}) and to a dendrogram, hence to a hierarchy of
segmentations. Efficient algorithms to produce saliency maps are the
subject of several studies. A basic algorithm (non-dedicated) can be
found in~\citep{najman2011equivalence}, but we rather recommend using
the more efficient one proposed in~\citep{cousty2011incremental}, with
the tree structure proposed in \citep{najman2013playing}.

\subsection{Design of criteria}
\label{sec:criterion}

In the previous sections \ref{sec:confilt} and \ref{sec:hiermst}, we
briefly describe several criteria.  In applications, the design of a
criterion adapted to the task at hand is fundamental. In mathematical
morphology, the first criteria proposed in the literature were of a
geometrical nature, such as the measure of the area
\citep{serra1992overview,vincent1994morphological} of the component,
or the depth \citep{grimaud1992new} or the volume
\citep{vachier1995extinction} of the blob corresponding to the
component. Stochastic criteria were developped in
\citep{angulo2007stochastic,meyer2010stochastic}, with an efficient
algorithm relying on watershed cuts in
\citep{malmberg2014efficient}. Optimisation of energy-type criteria
that make a balance between a data-attachment term and a
regularization term, were introduced latter
\citep{salembier2000binary}, and a formalization has been proposed
under the name of {\em scale-set} theory \citep{GuiguesCM06}.  A
recent review paper is available in \citep{salembier2009connected}.  A
generalization of the scale-set theory is proposed in
\citep{kiran2013global}.

Most of the previous criteria are increasing, allowing to transfrom
a hierarchy into another hierarchy. Non-increasing criteria are
frequent in the literature \citep{Zahn71,Morris1986,FH-04}, a simple
geometrical example being the various moments
\citep{westenberg2007volumetric}.  A generic framework for dealing with
non-increasing criteria has been proposed in
\citep{Xu2013,XGN-12,xu2013two}. Finally, we would like to mention
other approaches based on classical classification tools
\citep{guigues2003hierarchy} or on the {\em Helmotz principle}
(popularized in Computer Vision under the term {\em Number of False
  Alarms}) \citep{Cardelino2013}.






\section{A little further with graphs: discrete
  calculus}
\label{sec:further}



We have not reviewed in this paper numerous other interesting
graph-based approaches. Differential equations is one of them. Indeed,
discrete settings are recently becoming the subject of numerous
studies~\citep{grady2010discrete,desbrun2005discrete}: the main idea is
that one can write on graphs an exact discrete version of differential
equations, and efficiently solve many problems. For example, some
graph generalizations of the partial differential equations of
mathematical morphology~\citep{alvarez1993axioms} can be
written~\citep{ta2011nonlocal,drakopoulos2012active,purkait2012super},
offering a greater flexibility than the continuous framework (notably,
an easy integration of patch-based processing and novel applications).

\begin{figure*}[htbp]
  \begin{center}
    \begin{tabular*}{0.95\linewidth}{@{\extracolsep{\fill}}c c }
      \includegraphics[totalheight=0.43\textwidth]{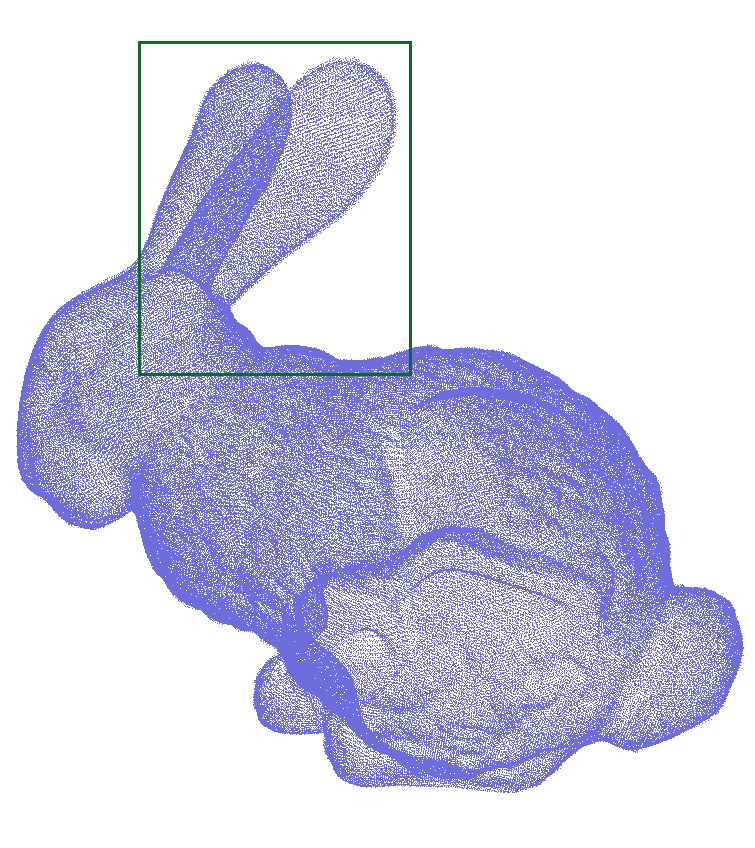}
      & \includegraphics[totalheight=0.43\textwidth]{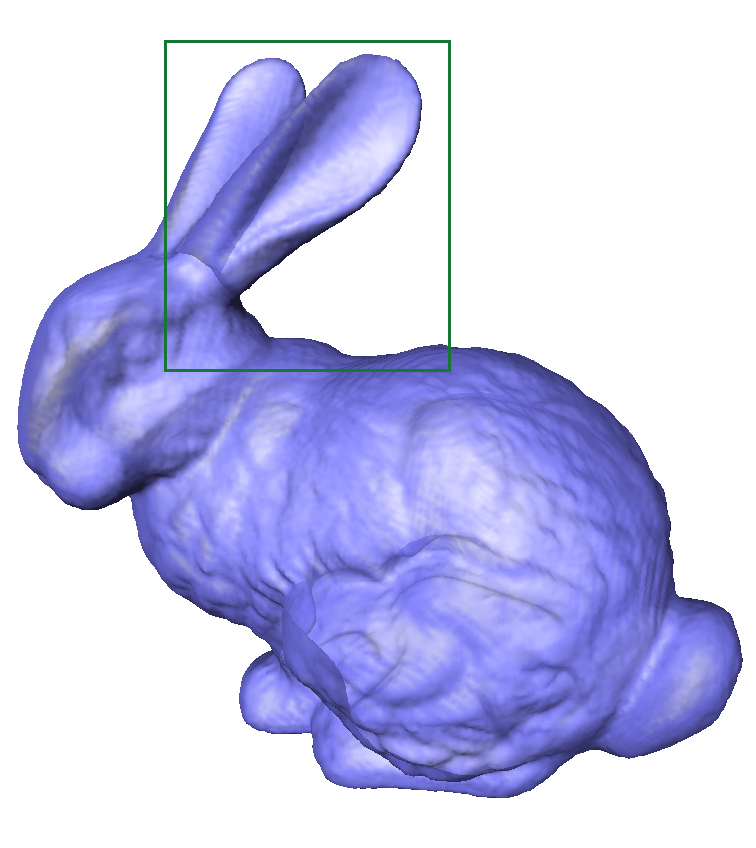}
      \\
      \includegraphics[totalheight=0.43\textwidth]{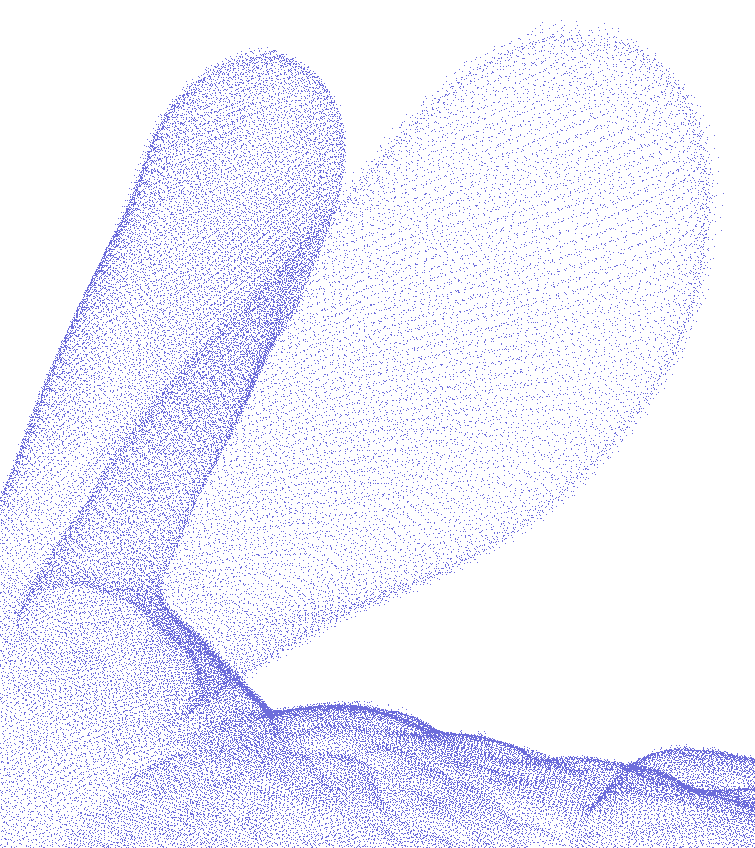}
      & \includegraphics[totalheight=0.43\textwidth]{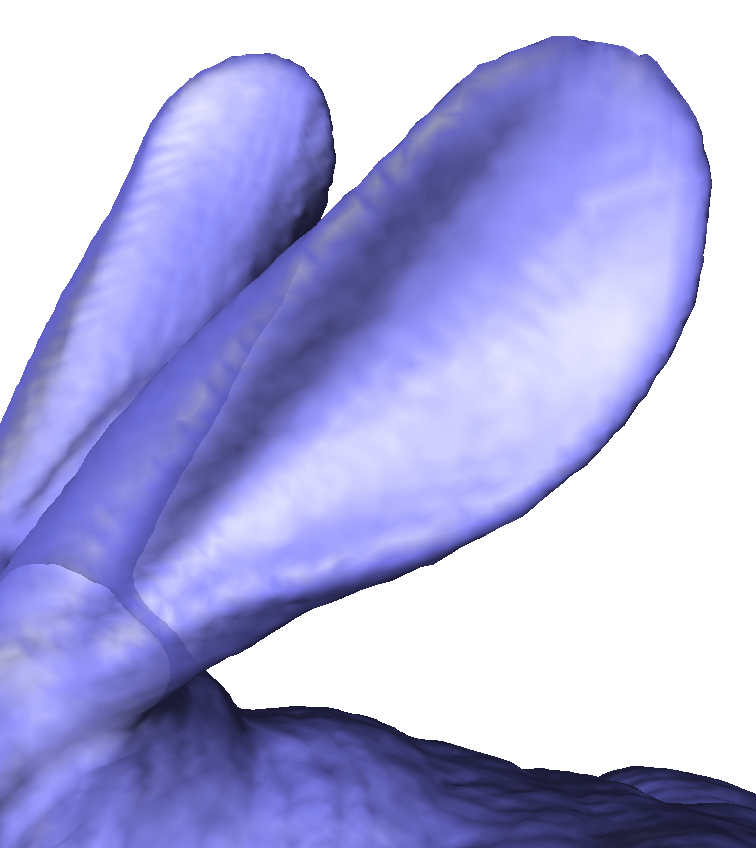}
    \end{tabular*}
  \end{center}
  \caption{\label{fig:surfrec} Surface reconstruction with the power
    watershed. From a noisy set of point measurements (left), a
    dedicated watershed algorithm with global optimality properties
    computes a smooth surface (right). The algorithm is fast, robust
    to seed placements, and compares favorably with existing
    algorithms~\citep{couprie2011surface}}
\end{figure*}

We would like to mention the popular graph-based optimization
approaches, such as the max-flow/min-cut
one~\citep{ford1962flows,LRCC-01} (known in the computer vision
community under the name of graph-cut~\citep{BVZ-01}). These methods
can be used to solve a wide variety of problems that can be formulated
in terms of energy minimization. Although energy minimization
approaches seem hardly related to the morphological approach based on
lattice theory~\citep{serra2006lattice}, there exists a framework
(called the power-watershed framework~\citep{CGNT-11}) in which
graph-cuts~\citep{BVZ-01}, shortest paths~\citep{falcao2004image},
random walks~\citep{grady2006random} and watershed cuts~\citep{CBNC-09},
can all be unified together, and in which we can study their links and
differences. Many applications can be designed thanks to this
framework, including some that are surprising for morphology: for
example the (power) watershed can now be used to perform the
anisotropic diffusion process~\citep{couprie2010anisotropic} or to
produce a surface reconstruction from unstructured cloud
points~\citep{couprie2011surface} (see Fig.~\ref{fig:surfrec}).

We believe that many other links with seemingly unrelated methods can
be searched and found: for example, the popular mean-shift
approach~\citep{cheng1995mean,comaniciu2002mean} can be
seen~\citep{paris2007topological} as computing a max-tree in the
feature space, and filtering this max-tree with a depth
criterion. Exploring, detailing and emphasizing such links with other
methods is indeed a promising research direction.

\section{Beyond graphs: other interesting structures}
\label{sec:beyond}

Several problems related to image processing cannot be handled with
undirected graphs as presented in this article.

The set of all connected sets of vertices in a graph form an algebraic
structure called a connection which was introduced in \citep[Chapter
2]{Serra-88} and further studied notably in
\citep{Ronse-95,BG-03,ronse2008partial}. The structure of a connection
is a basis for studying the algebraic properties related to
connectivity in many frameworks. Whereas the notion of a graph hardly
extends to the case of a continuous plane, a continuous setting can be
studied through a connection. Furthermore, even in the case of a
finite set of vertices, the notion of a connection is more versatile
than the one of a graph: for instance, with a connection, we can
consider the situation where a set of three points is connected
whereas any pair made of two of these three points is disconnected (in
a graph at least two of these three possible pairs must be connected
by an edge if the whole triple is connected). Such a connection could
be obtained using, for instance, an hypergraph.

A study of morphological operators on hypergraphs was recently
initiated by \citet{BB-13,BBL-13}. This framework
allows higher order information to be taken into account by grouping
any number of vertices into an hyper-edge. In particular, new
similarity measures between images were proposed based on
morphological operators in hypergraphs.

Asymmetric links between pairs of data cannot be considered in the
presented framework of undirected graphs. This information can be
taken into account in the framework of directed graphs. Image
processing, including in particular morphological processing, in this
kind of space is currently an emerging research topic
\citep{TTP-13,PCTTP-13,MM-13,Ronse14}.

For a complete topological characterization of geometrical objects,
graphs (as well as connections, hyper-graphs or directed graphs) are,
in general, not sufficient. Indeed, in a graph, we can make the
difference between a 0-dimensional element (a vertex) and a
1-dimensional element (an edge) but the distinction with a
2-dimensional element (\ie{} a patch of surface) cannot be made
without any further information. Moreover, whereas the ``cavities'' of
an object can be well identified with graphs as connected components
of the complement of an object, characterizing a hole such as the one
appearing in a torus is not feasible. Simplicial and cubical complexes
generalize graphs to higher dimensions in the sense that a graph is a
complex of dimension~1; furthermore, they allow the topological issues
mentioned above to be tackled
\citep{Bertrand-07cras,CouBer-09}. Intuitively, a simplicial complex
may be thought of as a set of elements having various dimension (\eg{}
tetraedra, triangles, edges, vertices) glued together according to
certain rules. Recent studies investigated mathematical morphology in
this framework, leading to morphological operators that can filter
noise with respect to its dimension \citep{DCN-11} and to links between
the notions of watershed and of homotopy
\citep{cousty:hal-00871498}. The framework of combinatorial maps, which
provides another topology-endowed representation of discrete objects,
has also been used to perform morphological filters of an image along
watershed contours before building a hierarchy of segmentation
\citep{BMM-05}.

\section{Conclusion}
As can be seen from this paper, graphs have been and currently are a
prominent topic in image analysis and computer vision. With the advent
of the so-called {\em Big Data}, we expect this trend to be extremely
persistent~\citep{lum2013extracting} and promising for opening novel
research directions. Indeed, there is no reason to restrict the
application of the very same ideas we have described here to
images. Any kind of data can be processed with these techniques,
notably, social graph models~\citep{grady2010discrete} (allowing
fine-graine prediction of human behavior), but also energy,
transportation, sensor and neural networks to name a few.

Most of the tools presented in this paper are readily available in
Pink, an open-source library~\citep{Pink,Pink-Marak}. In this library,
one can find various implementations of the very same operators,
according to the type of data (images, graphs, complexes, etc.) and
the value type (integer, float, color, etc.) that has to be
processed. A promising research direction is to write an algorithm
once, and let the compiler translate the resulting code to any type
of data one wants to deal with. This direction is pursued with the
Olena platform~\citep{Olena,levillain2009milena}, an open source
framework for generic data processing.





\section*{Acknowledgements}
\ELIMINE{The interest of the authors for the use of graphs in
  mathematical morphology was catalyzed from two concomitent
  events. In 2002, Laurent Najman joined ESIEE Paris as a teacher and
  researcher to form a group with Michel Couprie and Gilles Bertrand
  working in image processing. At the same time, Jean Cousty was an
  engineering student at ESIEE Paris where he attended the courses
  ``Graphs and algorithms'' and ``Initiation to mathematical
  morphology'' given by Gilles Bertrand and Michel Couprie. Shortly
  after, these four people started to work together on the so-called
  topological watershed which had been introduced, as a graph-based
  notion, by Gilles Bertrand and Michel Couprie in 1997.}  The work of
the authors on graph-based mathematical morphology owed much to the
collaboration with Gilles Bertrand and Michel Couprie and, therefore,
the authors feel indebted to them. The authors would also like to
thank Hugues Talbot and Christian Ronse for their careful reading of
the paper.






\bibliographystyle{model2-names}
\bibliography{SurveyMorphoGraph.bib}







\end{document}